\documentclass{article}
\usepackage{iclr2027_conference,times}
\usepackage{amsmath,amssymb}
\usepackage{booktabs}
\usepackage{longtable}
\usepackage{needspace}
\usepackage{graphicx}
\usepackage{float}
\usepackage[section]{placeins}
\usepackage{microtype}

\usepackage{multirow}
\usepackage{xcolor}
\usepackage{enumitem}
\usepackage{caption}
\usepackage[hidelinks]{hyperref}
\usepackage{url}
\usepackage{tikz}
\usepackage{algorithm}
\usepackage{algpseudocode}
\usetikzlibrary{arrows.meta,positioning,shapes.geometric}

\definecolor{panel}{HTML}{F1F3F6}
\definecolor{genfill}{HTML}{FFFFFF}
\newcommand{\eir}{EIR}

\title{Evidence-Inference Reconstruction: \\When The Evidence Is Recalled But The Reasoning Goes Wrong}
\author{Megan Diehl \& Ser-Nam Lim \\
University of Central Florida \\
\texttt{\{me419110, sernam\}@ucf.edu}}

\iclrfinalcopy

\begin{document}

\maketitle
\lhead{Preprint}

\begin{abstract}
Modern multi-hop LLM agents are equipped with built-in mechanisms to detect errors in intermediate reasoning steps. Such errors trigger corrective actions from these agents, which mostly follow the paradigm of retrying the steps or the reasoning trajectories. Not only are these retries expensive, we present in this paper that they are also potentially unnecessary. To this end, we introduce Evidence-Inference Reconstruction (EIR), which uses structured state to guide one retrieval trajectory, accumulating source evidence in the process. We show that as long as the relevant evidence has been collected, EIR is capable of generating the correct answer in a single final model call even if erroneous evidence has been mixed in due to incorrect intermediate reasoning steps.
In one evaluation, using Haiku 4.5 and GPT-4.1 Mini, we evaluate EIR on matched 1,000-question subsets of HotpotQA, 2WikiMultiHopQA, and MuSiQue, showing that EIR improves Answer F1, the overlap between the model's and the correct answer, over the baseline by 8.3--32.8 points, Agentic SSR by 10.6--29.1 points, and Reflexion by 1.1--15.9 points. Additionally, we show that EIR averages 4.85 total model calls per question, compared with 35.29 for Agentic SSR and 12.41 for Reflexion. Together, these results corroborate EIR's central premise: separating evidence retrieval from the final answer model call can improve answer accuracy while utilizing substantially less computation.
%
\end{abstract}

\begin{figure}[H]
\centering
\setlength{\abovecaptionskip}{3pt}
\setlength{\belowcaptionskip}{-4pt}

\begin{minipage}[t]{0.42\textwidth}
\vspace{0pt}
\centering
\textbf{(a) EIR architecture}

\vspace{2pt}

\includegraphics[
    width=0.82\linewidth,
    keepaspectratio
]{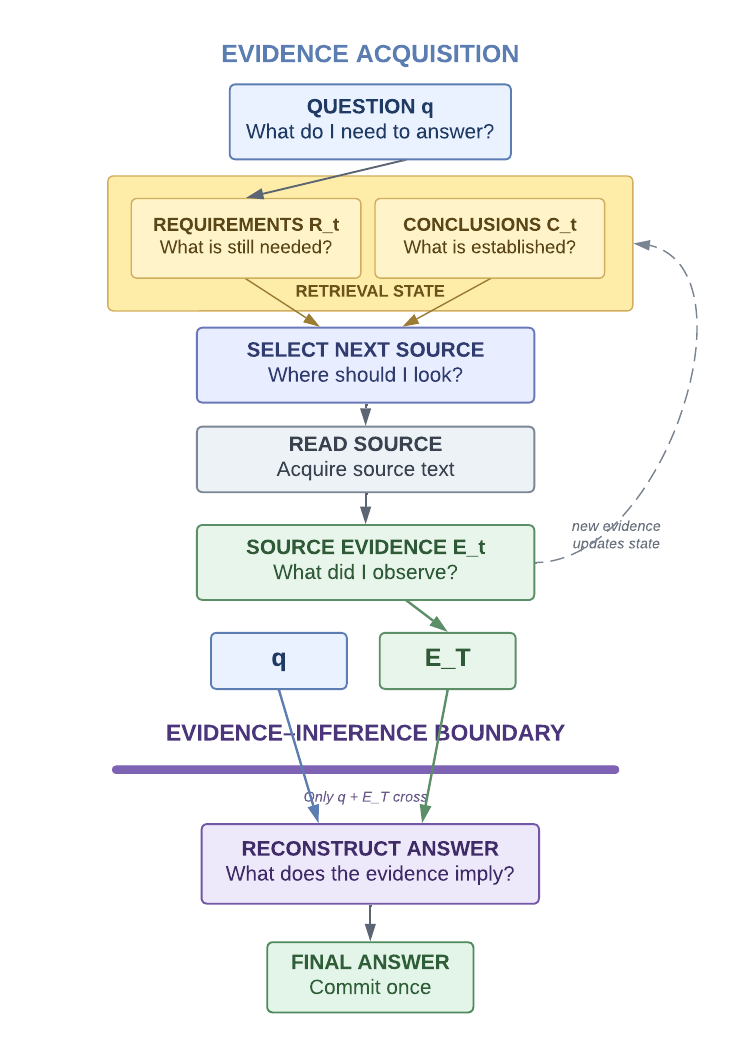}
\end{minipage}
\hfill
\begin{minipage}[t]{0.55\textwidth}
\vspace{0pt}
\centering

\refstepcounter{algorithm}
\label{alg:eir}
\textbf{(b) Algorithm \thealgorithm: Evidence-Inference Reconstruction}

\vspace{2pt}
\hrule
\vspace{2pt}

{\scriptsize
\begin{algorithmic}[1]
\Require Question $q$, paragraph titles $D$, read budget $H$
\Ensure Answer $\hat y$, supporting references $\hat F$
\State $E_0,R_0,C_0 \gets \emptyset$; $U_0 \gets D$
\State $o_0 \gets \emptyset$
\For{$t=0,\ldots,H$}
    \State $T \gets t$
    \State $(\Delta R_t,\Delta C_t,a_t)
        \gets \pi_\theta(q,o_t,R_t,C_t,U_t,H-t)$
    \If{the complete proposed list update is valid}
        \State Apply $(\Delta R_t,\Delta C_t)$ to obtain
            $R_{t+1},C_{t+1}$
    \Else
        \State $R_{t+1}\gets R_t$; $C_{t+1}\gets C_t$
    \EndIf
    \If{the submission is accepted or $t=H$}
        \State \textbf{break}
    \EndIf
    \State $E_{t+1}\gets E_t$; $U_{t+1}\gets U_t$;
        $o_{t+1}\gets o_t$
    \If{$a_t=\mathrm{READ}(d_t)$ and $U_t\neq\emptyset$}
        \State If $d_t\notin U_t$, use the first unread title
        \State $o_{t+1}\gets x_{d_t}$
        \State $E_{t+1}\gets E_t
            \cup\{(d_t,i,s_{d_t,i})\}_{i=1}^{|d_t|}$
        \State $U_{t+1}\gets U_t\setminus\{d_t\}$
    \EndIf
\EndFor
\State $(\hat y,\hat F)\gets f_{\mathrm{boundary}}(q,E_T)$
\State \Return $(\hat y,\hat F)$
\end{algorithmic}
}

\vspace{2pt}
\hrule
\end{minipage}

\vspace{3pt}
\caption{\textbf{Evidence-Inference Reconstruction.}
\textbf{(a)} Requirements and evidence-citing conclusions guide document selection while retrieved source evidence is accumulated separately.
\textbf{(b)} Retrieval ends with one final reconstruction from only the original question $q$ and accumulated evidence $E_T$; the retrieval state and provisional answer are withheld.}
\label{fig:eir-overview}
\vspace{-6pt}
\end{figure}

\section{Introduction}

LLM agents reason within an action-state loop: each action changes what the agent observes, and each observation allows the agent to reason about what to do next~\citep{yao2023react}. An error that occurs in an intermediate reasoning step could cascade, potentially causing the agent to encounter evidence that could be erroneous in subsequent steps. State-of-the-art methods like Socratic Self-Refine (SSR) \citep{shi2025ssr} and Reflexion \citep{shinn2023reflexion} have built-in mechanisms to detect these errors, from which they try to recover by revisiting the reasoning itself. SSR, particularly our agentic adaptation of it, detects such an error when the intermediate reasoning step is associated with a low confidence score, and then proposes corrective actions to repeat the current step. Reflexion, on the other hand, diagnoses unsuccessful trajectories at the end based on the difference between the human-annotated ground-truth label and the trajectory output, and then uses this difference to guide subsequent attempts. Intuitively, both methods follow the paradigm of rectifying the detected errors, repeatedly if necessary, in the hope of converging towards a correct final answer.

Two observations can be made here. First, approaches like these, while intuitive, incur substantial computational costs due to the retries. Second, we observe that even along an incorrect reasoning trace, key \emph{correct} evidence often has already been discovered, even though it is mixed in with the erroneous ones. Riding on these observations, we propose Evidence-Inference Reconstruction (EIR), whose central idea is that one does not have to conduct any intermediate or trajectory retries as long as the key evidence needed for the final correct answer has already been discovered in the reasoning trace. That is, as long as the right evidence is already present, a single final model inference call with the collected evidence can yield the correct answer, saving precious computation. In particular, EIR uses requirements and conclusions to guide one retrieval trajectory, storing the retrieved source text separately, then answering in a single final model call. Retrieval still influences the answer through the evidence selected, but it's the final inference call that ultimately produces the answer. Preserving the source text allows reconsideration without unnecessarily repeating the actions that acquired it.

\begin{figure}[!t]
\centering
\includegraphics[
    width=\textwidth,
    keepaspectratio
]{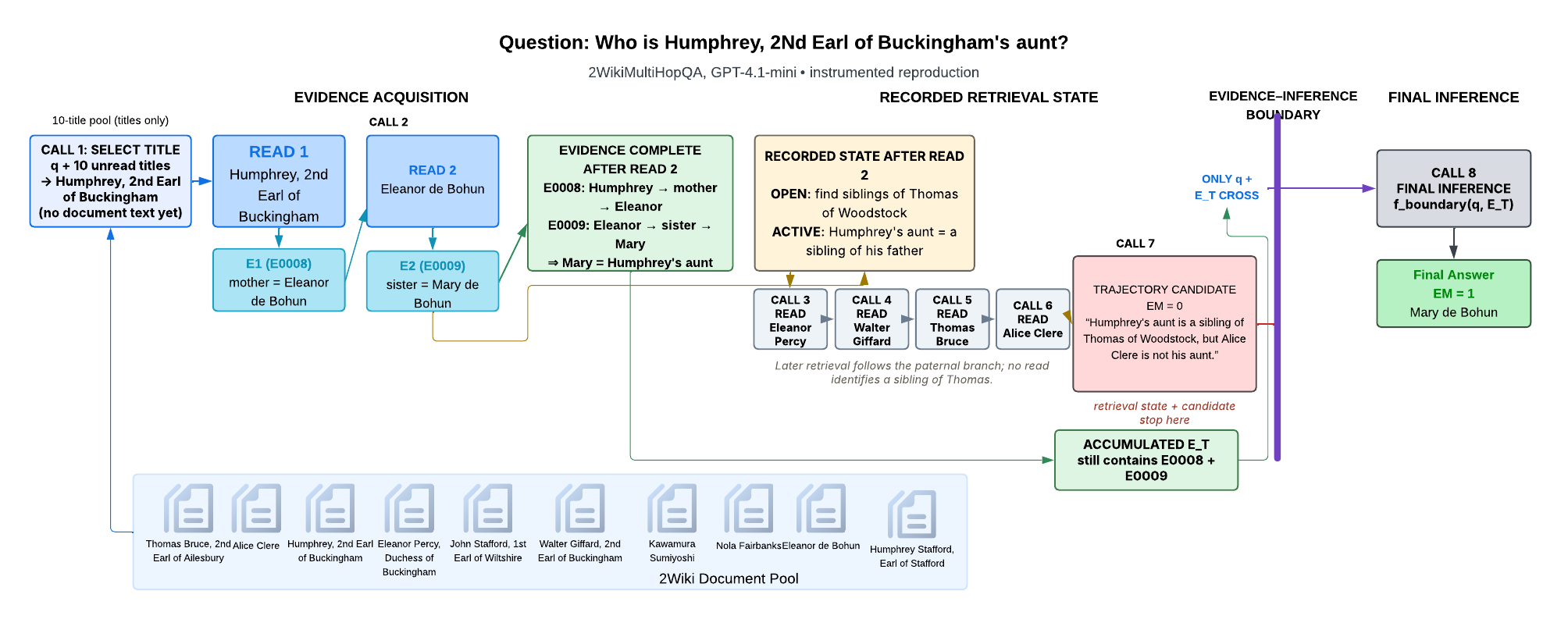}
\caption{\textbf{EIR trace showing an incorrect intermediate candidate followed by a correct final answer.}
This instrumented reproduction, outside the confirmatory protocol, produces an incorrect candidate despite retrieving the necessary evidence. A fresh call over only the question and source text derives the correct answer. Appendix~\ref{app:trace-comparison} documents its eight-call procedure.}
\label{fig:eir-recorded-trace}
\end{figure}

These observations can be seen in the Humphrey example below (Figure~\ref{fig:eir-recorded-trace}). Both Agentic SSR and Reflexion retrieve the evidence that identifies Eleanor de Bohun as Humphrey's mother and Mary de Bohun as Eleanor's sister. Given this information, both models should already be able to answer the question, "Who is Humphrey, 2nd Earl of Buckingham's aunt?", using the connection that an aunt is the sister of a mother. Yet, Agentic SSR continues to look into further branches, ultimately answering with a woman completely unrelated to Humphrey (See Appendix~\ref{app:trace-comparison}, Figure~\ref{fig:ssr-recorded-trace}). In the same vein, Reflexion repeatedly searches for Humphrey’s parents and their siblings despite already having sufficient evidence, then misidentifies his mother as his aunt in all five trials (see Appendix~\ref{app:trace-comparison}, Figure~\ref{fig:reflexion-recorded-trace}). Moreover, both methods incur substantial computation while retaining incorrect answers: Agentic SSR makes 66 model calls within one trajectory, while Reflexion makes 22 calls across five trials, including four repeated attempts. Other Reflexion runs exhaust their turn or retry budgets and ultimately produce erroneous conclusions (See Appendix~\ref{app:corpus-diagnostics}). EIR, by contrast, does not rely on intermediate reasoning to form its final answer except through the evidence that reasoning selects, allowing it to correctly identify Mary de Bohun as Humphrey's aunt after acquiring all the evidence (Figure~\ref{fig:eir-recorded-trace}). As a result, it substantially cuts down the total number of model calls while largely producing better accuracy scores across all datasets.

In particular, we show in our benchmark evaluations that EIR averages 4.85 model calls per question, whereas Agentic SSR averages 35.29, and Reflexion 12.41. In addition, EIR achieves a higher average Answer F1 than both methods, a metric of overlap between the model's answer and the correct answer, as provided by the HotpotQA \citep{yang2018hotpotqa}, 2WikiMultiHopQA \citep{ho2020twowiki}, and MuSiQue \citep{trivedi2022musique} datasets. Specifically, using GPT-4.1 Mini \citep{openai2025gpt41} and Haiku 4.5 \citep{anthropic2025haiku45} on matched 1,000-question subsets of these benchmarks, we show that EIR improves Answer F1 by 8.3--32.8 points over the baseline, 10.6--29.1 over Agentic SSR, and 1.1--15.9 over Reflexion (Table~\ref{tab:main}). Together, these results indicate that an agent does not need to revisit or correct its intermediate reasoning to produce an evidence-backed answer to multi-hop questions. 
%
%

In sum, our main contributions are twofold:
    (1) We introduce \textbf{Evidence-Inference Reconstruction (EIR)}, a novel agentic paradigm that retrieves evidence in a single trajectory and then generates the final answer in a separate inference step using the collected evidence. This separation of evidence collection and final inference effectively avoids additional model calls that corrective reasoning methods can introduce; and
%
    (2) EIR outperforms its counterparts overwhelmingly. Our matched comparison evaluations, using GPT-4.1 Mini and Haiku 4.5, show that EIR significantly improves Answer F1 over Agentic SSR in all six settings and over oracle-stopped Reflexion in five of six and uses fewer average model calls than either method, reducing overall test-time computation.
%

\section{Related Work}
Recent work has attempted to address these limitations of corrective reasoning paradigms, but often in ways that remain computationally expensive and unnecessarily complex. In particular, there is a shift toward agentic Retrieval-Augmented Generation (RAG) methods. Such methods integrate real-time data retrieval from external sources into an agentic environment \citep{asai2024selfrag,jeong2024adaptiverag}. However, such methods often share the same underlying issue as corrective reasoning methods like Socratic Self-Refine \citep{shi2025ssr} and Reflexion \citep{shinn2023reflexion}: an incorrect intermediate reasoning step can cascade and potentially lead to a wrong final answer. Other methods attempt to store the evidence alongside the reasoning \citep{liu2026sleuth} or track evidence gaps alongside the findings \citep{asl2025fairrag}, but they do not separate the evidence retrieval entirely from the final answer. As such, potentially incorrect intermediate reasoning can still cause the model to answer incorrectly. EIR, on the other hand, answers in a single final model call, using only the final accumulated evidence. As a result, EIR can answer correctly even if the collected evidence contains information irrelevant to the question, as long as the relevant information has been retrieved.

\section{Method}
\label{sec:method}

The methods above primarily either revisit their reasoning when it goes wrong or carry it into the final answer. EIR takes a different approach, building on the observation that an agent does not need to waste further computation once it has already retrieved the evidence needed to answer the question. The agent can instead answer in a single model call from the accumulated evidence alone. This is the intuition behind EIR: during evidence retrieval, requirements identify potential missing information, and conclusions record claims the agent considers supported. After each document is read, the agent reasons about what key information is still needed to answer the question, then it updates the requirements and conclusions to determine the next document to read.

Once retrieval ends, either because the agent submits an answer or because it exhausts its retrieval budget, a separate final model call answers from only the original question and the retrieved source text. As a result, the reasoning written during retrieval, including any mistakes in it, does not reach the final answer directly. Both the evidence acquisition and the final model call use the same language model without additional training. Sections~\ref{sec:task}--\ref{sec:reconstruction} define the task, the retrieval state, and the final answer call, and Algorithm~\ref{alg:eir} summarizes the procedure.

\subsection{Task and Interaction Model}
\label{sec:task}

Before describing each part of this paradigm, we first define the setting in which the agent answers a question. We refer to the full process of answering one question as an \emph{episode}. An episode begins with a question \(q\), the titles \(D\) of a pool of paragraphs that the benchmark supplies with that question, and a turn budget \(H\). A \emph{turn} is one step in which the agent takes a single action, and the \emph{turn budget} \(H\) is the maximum number of turns the agent is allowed for reading during retrieval, followed by an additional turn for the final model call. At the start of the episode, the agent sees only the question and the titles.

Because the agent sees a paragraph's text only after reading it, the procedure tracks three quantities at each turn \(t\): the \emph{unread titles} \(U_t\subseteq D\), the \emph{current observation} \(o_t\), which is the text of the most recently read paragraph, and the \emph{evidence store} \(E_t\), which holds every sentence read so far. Initially, \(U_0=D\) and \(o_0=E_0=\emptyset\), then at each turn, the agent either reads an unread paragraph or submits an answer,
\[
\mathcal{A}_t=
\{\mathrm{READ}(d):d\in U_t\}
\cup
\{\mathrm{SUBMIT}(y,F)\},
\]
where \(y\) is the submitted answer, and \(F\) the list of sentences the agent cites as supporting evidence. When the agent reads a paragraph \(d_t\), with text \(x_{d_t}\), \(i\)-th sentence \(s_{d_t,i}\), and \(|d_t|\) sentences, all three quantities update:
\[
\begin{aligned}
o_{t+1}&=x_{d_t},\qquad U_{t+1}=U_t\setminus\{d_t\},\\
E_{t+1}&=E_t\cup\{(d_t,i,s_{d_t,i})\}_{i=1}^{|d_t|}.
\end{aligned}
\]
The paragraph's text \(x_{d_t}\) becomes the new observation \(o_{t+1}\), which the agent uses to choose its next action, and EIR removes the read paragraph \(d_t\) from the list of unread titles \(U_t\), so that no paragraph is read twice. Each sentence of the paragraph \(d_t\) also enters the evidence store as a \emph{record}, which holds the paragraph title, the sentence's position within the paragraph, and the sentence text. EIR builds these records directly from the retrieved paragraph, so the model can cite a record but cannot create or alter one (Appendix~\ref{app:impl}). While each new read replaces the observation, the evidence store keeps every record. By the end of retrieval, the accumulated evidence \(E_T\) therefore holds an exact copy of every sentence the agent has read, independent of how the agent interprets it.

\subsection{State-Guided Evidence Acquisition}
\label{sec:acquisition}

The evidence store records what the agent has read, but not what the agent has concluded from it or what it still needs to find. EIR keeps this interpretation in two lists that the agent carries from one turn to the next. The \emph{requirement list}
\[
R_t=[r_{t,1},\ldots,r_{t,m_t}],
\qquad m_t=|R_t|,
\]
holds the information the agent still needs, where each requirement \(r_{t,j}\) describes one missing entity, relation, attribute, or comparison. The \emph{conclusion list}
\[
C_t=[(c_{t,j},S_{t,j})]_{j=1}^{n_t},
\qquad n_t=|C_t|,
\qquad S_{t,j}\subseteq E_t,
\]
holds the claims the agent currently considers supported, pairing each claim \(c_{t,j}\) with the records \(S_{t,j}\) cited for it. Here, \(m_t\) and \(n_t\) are the numbers of requirements and conclusions at turn \(t\). Each requirement or conclusion, which we call an \emph{entry}, also has an identifier and a status that the model uses to refer to and revise it.

At each turn, a single model call receives the question, the current observation, both lists, the unread titles, and the number of turns remaining, and it returns changes to both lists along with the next action:
\[
(\Delta R_t,\Delta C_t,a_t)
\gets
\pi_\theta(q,o_t,R_t,C_t,U_t,H-t),
\]
where \(\Delta R_t\) and \(\Delta C_t\) are the list changes and \(a_t\in\mathcal{A}_t\). We refer to this call, \(\pi_\theta\), as the \emph{retrieval policy}. Because the same call produces both the list changes and the action, maintaining the lists adds no model calls. The lists start empty, so the first action depends only on the question and the titles. Afterward, the lists and the current observation inform every action, including the decision to submit, so the lists connect what the agent has concluded to what it reads next.

Before applying any change, EIR checks that it revises only existing entries and cites only records already in \(E_t\), then discards any change that fails this check (Appendix~\ref{app:impl}). This check confirms that the cited records were read, not that they support the claim. As a result, the lists can still contain mistakes. A wrong conclusion can send the agent toward the wrong documents, and an answer built on it can be wrong even when \(E_t\) holds the sentences needed to answer correctly.

\subsection{Final Answer Reconstruction}
\label{sec:reconstruction}

To prevent incorrect intermediate reasoning from altering the final answer, EIR separates retrieval from answering. Retrieval ends at turn \(T\leq H\), either after the agent submits or after it reaches the turn budget. A submission
\[
a_T=\mathrm{SUBMIT}(\tilde{y}_T,\tilde{F}_T)
\]
contains the answer \(\tilde{y}_T\) that the agent proposes during retrieval and the sentences \(\tilde{F}_T\) it cites. EIR uses this submission only to end retrieval. Instead of scoring \(\tilde{y}_T\), it produces the scored answer with a new call to the same language model under a separate final-answer prompt:
\[
(\hat{y},\hat{F})=f_{\mathrm{boundary}}(q,E_T),
\]
where \(\hat{y}\) is the final answer, and \(\hat{F}\) the list of sentences it cites. We refer to \(f_{\mathrm{boundary}}\) as the final answer call, or the final model call, in other parts of the paper. It runs once retrieval ends, including when the agent exhausts its turn budget without submitting.

What distinguishes the final answer call from the retrieval policy is its input: it receives only the question \(q\) and the records in the accumulated evidence \(E_T\). We call this restriction the \emph{evidence--inference boundary} because the evidence crosses it, while the inferences the agent made to retrieve the evidence do not. The agent's reasoning therefore shapes the final answer only through which paragraphs it selected to read.

This restriction allows EIR to recover from reasoning errors without retrying. A misinterpretation formed during evidence retrieval is not inherited in the final answer (Figure~\ref{fig:eir-recorded-trace}), however, by the same restriction, the final answer call also cannot recover information that the agent never discovered (Appendix~\ref{app:failure-analysis}).

\section{Experiment Design}
\label{sec:experiment-design}


\subsection{Evaluation Setup}
\label{sec:setup}

We evaluate EIR, Agentic SSR, Reflexion, and all ablations and conditions across HotpotQA \citep{yang2018hotpotqa}, 2WikiMultiHopQA \citep{ho2020twowiki}, and the answerable portion of MuSiQue \citep{trivedi2022musique}. For every question supplied by the benchmarks, there is a corresponding set of paragraphs that either do or do not support the expected answer. The number of paragraphs per question varies by the evaluation protocol dictated by the benchmark: ten for HotpotQA and 2WikiMultiHopQA, and twenty for MuSiQue. In total, there are 1,000 questions per benchmark; therefore, HotpotQA and 2WikiMultiHopQA have a total of 10,000 paragraphs, and MuSiQue has a total of 20,000. We evaluate every method using these same 1,000 questions, using Haiku 4.5 \citep{anthropic2025haiku45} and GPT-4.1 Mini \citep{openai2025gpt41}. Appendix~\ref{app:repro} describes the dataset files, question selection, and fixed model versions.

\subsection{Baselines}
\label{sec:conditions}

Our baseline, \emph{Direct}, interleaves reasoning and paragraph selection \citep{yao2023react}, and we evaluate its first valid submitted answer. EIR instead maintains a list of requirements and conclusions while reading and selecting paragraphs, which determines whether to continue reading, including which paragraphs to read, or to submit once there is sufficient evidence to answer. \emph{Direct} has a maximum of six turns to execute all actions, including reading and submitting its answer (\(H=6\)), whereas EIR has a maximum of six turns to read, and a single final turn to submit. \emph{Direct} and EIR both use one model call for each read action; however, EIR also uses an extra model call for its final answer generation.

We also consider what we referred to as \emph{Full Pool}, which tests whether EIR benefits from its selective evidence retrieval mechanism. It allows the model to simply read every paragraph before answering the question instead of selecting only certain relevant paragraphs. However, this is clearly impractical in the real world in which the data volume is significantly larger than a single paragraph per document, and there are potentially millions of documents to read from. Therefore, \emph{Full Pool} is primarily an ideal representation of EIR, which helps to shed useful insights in our comparisons. All remaining controls and their results appear in Appendices~\ref{app:mechanism-ablations} and~\ref{app:failure-analysis}.



\emph{Agentic SSR} adapts Socratic Self-Refine \citep{shi2025ssr} to examine each proposed action before executing it. After the model proposes an action, a separate call decomposes the proposal into three reasoning steps: identifying established facts, explaining how those facts justify the proposed action, and specifying the next action to execute. Each step then receives three verifier judgments using a model temperature of 0.7, and it qualifies for refinement if the majority of the judgments vote for a correction. If any step qualifies, an additional call refines the step with the lowest confidence score before executing the proposed action. Otherwise, refinement does not occur. This verification results in significant computation even when no refinement occurs, as it always executes.

\emph{Reflexion} \citep{shinn2023reflexion} repeats retrieval for up to five trials, carrying written feedback from unsuccessful trials into the subsequent ones. Both Agentic SSR and Reflexion choose from the same paragraphs as Direct and EIR, with a maximum of six turns per retrieval trajectory. \emph{Reflexion first trial} reports the initial answer before any retries, showing the performance without those additional calls. \emph{Oracle-stopped Reflexion} reports the first answer that exactly matches the reference answer from the dataset, or the final trial's if none matches. The term \emph{oracle} refers to access to information that would normally be unavailable: the correct answer to the question. In \emph{oracle-stopped Reflexion}, an external controller uses the benchmark’s reference answer to stop at the first exact match or allow another trial, up to five trials. The model itself never receives the reference answer. Reporting both shows the improvement obtained from repeated attempts together with this stopping rule. Appendix~\ref{app:comparison-adaptation} further details these adaptations.

\subsection{Extended Evaluation with Corpus Search}
\label{sec:corpus-design}
To test transfer to a setting where relevant documents are not provided in advance, an additional experiment uses the same 1,000 HotpotQA \citep{yang2018hotpotqa} questions and both language-model backbones but searches through 5.23 million Wikipedia introductory paragraphs. This expanded paragraph collection is provided by HotpotQA. Here, the agent first supplies search terms through a \textsc{Search} action, and a keyword-ranking method, called BM25 ($k_1=0.9$, $b=0.4$), gives more weight to rare search terms and adjusts for paragraph length, returning ten or fewer titles and unique document identifiers. A \textsc{Read} action then uses one of these identifiers to obtain the paragraph's text. EIR's requirement and conclusion lists guide \textsc{Search} and \textsc{Read} actions, as they guide \textsc{Read} and \textsc{Submit} in the standard evaluation. Each trajectory permits a maximum of 12 action turns, including submission, and six maximum read attempts per turn. All calls use a model temperature of 0.0 and a 2,048-token limit, including SSR verification. Appendix~\ref{app:corpus} specifies the corpus and search interface further.

\subsection{Metrics and Statistical Procedure}
\label{sec:stats}

\emph{Answer F1}, our primary metric, measures word overlap with the reference answer. Alongside it, we report \emph{Exact Match} (EM), which requires a match after benchmark normalization, and \emph{Support F1}, which compares cited evidence with benchmark annotations. All three metrics are averaged over the 1,000 questions, including those with failed outputs, and reported on a 0--100 scale. To assess the computation required for these answers, we count model calls for action selection, verification, refinement, reflection, and final answering, excluding provider retries. These totals include Direct's calls, all proposal, decomposition, verification, and refinement calls for Agentic SSR, and all trial and reflection calls for oracle-stopped Reflexion.

Each comparison uses the two methods' scores on the same questions. For Answer F1, we resample these questions 10,000 times and use the middle 95\% of the resulting average differences as the confidence interval. For EM, an exact two-sided McNemar test compares the questions answered correctly by only one method. Holm correction reduces the risk of declaring differences by chance across multiple comparisons, applied separately to the groups defined in Appendices~\ref{app:mechanism-ablations}, \ref{app:comparison}, \ref{app:fullpool}, and~\ref{app:corpus}.

\section{Results}

\paragraph{Overall Performance.}
EIR improves Answer F1 over Direct by 8.27--32.79 points across the three benchmarks using both models, with every gain significant (Table~\ref{tab:main}, Panel A). EM and Support F1 also increase (See Appendix~\ref{app:confirmatory-results}, Table~\ref{tab:secondary}). Because Direct more often returns outputs that cannot be parsed as answers, we include only questions with readable outputs from both methods. EIR's Answer F1 gains remain significant at 3.76--30.65 points (See Appendix~\ref{app:parse}). The following comparisons examine the computation behind these gains and the contributions of evidence access, retrieval lists, and final answering.

\begin{table}[!t]
\centering\small
\setlength{\tabcolsep}{3pt}
\resizebox{\textwidth}{!}{%
\begin{tabular}{llrrrrrrr}
\toprule
\multicolumn{9}{l}{\textbf{Panel A. EIR and Direct}} \\
\midrule
 & & \multicolumn{2}{c}{Answer F1} & \multicolumn{3}{c}{EIR minus Direct} & \multicolumn{2}{c}{Calls/question} \\
\cmidrule(lr){3-4}\cmidrule(lr){5-7}\cmidrule(lr){8-9}
Model & Benchmark & Direct & EIR & Difference & \multicolumn{2}{c}{95\% interval} & Direct & EIR \\
\midrule
\multirow{3}{*}{Haiku 4.5}
 & HotpotQA & 64.37 & \textbf{72.64} & $+8.27$ & \multicolumn{2}{c}{[$+6.39$, $+10.18$]} & \textbf{2.85} & 4.52 \\
 & 2WikiMultiHopQA & 49.69 & \textbf{70.22} & $+20.53$ & \multicolumn{2}{c}{[$+17.95$, $+23.10$]} & \textbf{3.68} & 5.11 \\
 & MuSiQue & 40.82 & \textbf{51.98} & $+11.16$ & \multicolumn{2}{c}{[$+8.73$, $+13.66$]} & \textbf{3.49} & 5.32 \\
\midrule
\multirow{3}{*}{GPT-4.1 Mini}
 & HotpotQA & 56.19 & \textbf{74.05} & $+17.85$ & \multicolumn{2}{c}{[$+15.53$, $+20.13$]} & \textbf{2.78} & 4.08 \\
 & 2WikiMultiHopQA & 42.10 & \textbf{74.89} & $+32.79$ & \multicolumn{2}{c}{[$+30.13$, $+35.44$]} & \textbf{3.55} & 4.86 \\
 & MuSiQue & 34.42 & \textbf{57.54} & $+23.12$ & \multicolumn{2}{c}{[$+20.58$, $+25.70$]} & \textbf{3.09} & 5.19 \\
\midrule
\multicolumn{9}{l}{\textbf{Panel B. EIR and methods that verify actions or repeat retrieval}} \\
\midrule
 & & \multicolumn{4}{c}{Answer F1} & \multicolumn{3}{c}{Calls/question} \\
\cmidrule(lr){3-6}\cmidrule(lr){7-9}
Model & Benchmark & EIR & \shortstack{Agentic\\SSR} & \shortstack{Reflexion\\first trial} & \shortstack{Reflexion\\oracle stopping} & EIR & \shortstack{Agentic\\SSR} & \shortstack{Reflexion\\oracle stopping} \\
\midrule
\multirow{3}{*}{Haiku 4.5}
 & HotpotQA & \textbf{72.64} & 60.05$^{*}$ & 61.50$^{\dagger}$ & 67.66$^{\ddagger}$ & \textbf{4.52} & 31.34 & 10.90 \\
 & 2WikiMultiHopQA & \textbf{70.22} & 41.09$^{*}$ & 49.04$^{\dagger}$ & 65.95$^{\ddagger}$ & \textbf{5.11} & 40.30 & 12.59 \\
 & MuSiQue & \textbf{51.98} & 41.42$^{*}$ & 45.95$^{\dagger}$ & 50.86 & \textbf{5.32} & 38.43 & 12.77 \\
\midrule
\multirow{3}{*}{GPT-4.1 Mini}
 & HotpotQA & \textbf{74.05} & 59.50$^{*}$ & 61.47$^{\dagger}$ & 65.13$^{\ddagger}$ & \textbf{4.08} & 30.35 & 11.31 \\
 & 2WikiMultiHopQA & \textbf{74.89} & 48.23$^{*}$ & 55.93$^{\dagger}$ & 58.97$^{\ddagger}$ & \textbf{4.86} & 37.98 & 12.75 \\
 & MuSiQue & \textbf{57.54} & 34.83$^{*}$ & 37.65$^{\dagger}$ & 42.80$^{\ddagger}$ & \textbf{5.19} & 33.31 & 14.15 \\
\bottomrule
\end{tabular}}
\caption{\textbf{Main evaluation using the paragraphs provided with each question.} Each row represents 1,000 questions. Bold marks the highest Answer F1 and fewest calls within each displayed group. Reflexion's first trial includes no retry. Oracle stopping permits five trials and stops at the first exact match to the reference, with all trials and retries counted. Superscripts mark comparison scores significantly below EIR after Holm correction: Agentic SSR ($^{*}$), Reflexion first trial ($^{\dagger}$), and Reflexion with oracle stopping ($^{\ddagger}$).}
\label{tab:main}
\end{table}

\paragraph{Accuracy and Computation against Iterative Correction.}
\label{sec:comparison-results}
EIR exceeds Agentic SSR by 10.56--29.12 Answer F1 points and Reflexion's first trial by 6.04--21.18 points, significantly in every benchmark and model combination (Table~\ref{tab:main}, Panel B). Allowing Reflexion to repeat retrieval for up to five trials (oracle stopping) narrows the differences, but EIR retains significant gains in five of six comparisons. On MuSiQue using Haiku 4.5, the remaining 1.13-point difference is inconclusive, while EM favors Reflexion by 3.90 points (See Appendix~\ref{app:comparison}). \textcolor{black}{The 95\% confidence interval for EIR's Answer F1 advantage is  $[-1.36, +3.62]$ points.} Across the six comparisons, EIR averages 4.85 calls per question, compared with 35.29 for Agentic SSR and 12.41 for oracle-stopped (five trial) Reflexion. Its accuracy gains therefore accompany fewer calls than either complete correction procedure (See Appendix~\ref{app:comparison}, Table~\ref{tab:computation}).

\paragraph{Answering from All Provided Paragraphs.}
Full Pool answers in one call after reading all the given paragraphs, whereas EIR spends calls choosing what to read before answering from its selection. Despite Full Pool's complete evidence access, EIR significantly improves Answer F1 on 2WikiMultiHopQA by 5.60 points using Haiku 4.5 and 6.23 using GPT-4.1 Mini (Table~\ref{tab:simple}). Full Pool exceeds EIR on HotpotQA using GPT-4.1 Mini by 3.08 points, while the other three differences remain inconclusive with a 95\% confidence interval for EIR minus full pool being $[-1.30, +2.30$] points on HotpotQA using Haiku 4.5 . (See Appendix~\ref{app:fullpool}, Table~\ref{tab:fullpool-paired}).

Full Pool uses 6,000 calls compared with EIR's 29,082 because it eliminates calls to select paragraphs. EIR's 2WikiMultiHopQA gains show that Full Pool is a reference for complete evidence access rather than an accuracy ceiling. Its one-call procedure is practical for these small collections, but placing millions of paragraphs in one prompt would exceed the models' input capacity and would be impractical for most real-world cases. The corpus extension therefore tests the efficacy of EIR searching for the evidence among millions of candidate paragraphs.

\begin{table}[!t]
\centering\small
\begin{tabular*}{\textwidth}{@{\extracolsep{\fill}}llrr@{}}
\toprule
Model & Benchmark & Full Pool & EIR \\
\midrule
Haiku 4.5 & HotpotQA & 72.11 & \textbf{72.64} \\
 & 2WikiMultiHopQA & 64.62 & \textbf{70.22} \\
 & MuSiQue & \textbf{53.51} & 51.98 \\
\midrule
GPT-4.1 Mini & HotpotQA & \textbf{77.12} & 74.05 \\
 & 2WikiMultiHopQA & 68.66 & \textbf{74.89} \\
 & MuSiQue & \textbf{57.84} & 57.54 \\
\bottomrule
\end{tabular*}
\caption{\textbf{Answer F1 with every provided paragraph versus EIR's selected evidence.} Full Pool reads all paragraph text together and answers in one call, without model calls to select evidence. Bold marks the higher observed score, not statistical significance. Paired tests and intervals appear in Appendix~\ref{app:fullpool}, Table~\ref{tab:fullpool-paired}.}
\label{tab:simple}
\end{table}

\paragraph{Requirements and Conclusions during Retrieval.}
To test the requirement and conclusion lists' contribution, a separate control updates them but leaves them out of later retrieval prompts. Including the lists improves Answer F1 on 2WikiMultiHopQA by 0.80 points using Haiku 4.5 and 2.45 using GPT-4.1 Mini, and on MuSiQue by 4.83 and 1.30 points, respectively. The 2.45-point and 4.83-point gains are significant after Holm correction, while HotpotQA shows no significant benefit using either model (See Appendix~\ref{app:listfeedback}). A further variant revises conclusions when their supporting claims change, producing mixed effects relative to the default lists (See Appendix~\ref{app:repair}).

\paragraph{Transfer to Corpus Search.}
In the 5.23 million Wikipedia corpus evaluation, EIR retains significant Answer F1 gains over Direct, Agentic SSR, and Reflexion's first trial using both models (Table~\ref{tab:corpus-main}). It also exceeds oracle-stopped Reflexion by 4.68 points using GPT-4.1 Mini. Using Haiku 4.5, EIR achieves higher Answer F1 than oracle-stopped Reflexion on 199 of the 1,000 questions, lower Answer F1 on 170, and equal scores on 631. Because its losses are larger than its gains on average, its overall Answer F1 is 0.86 points lower. On the other hand, EIR requires significantly much fewer calls than Agentic SSR and oracle-stopped Reflexion (See Appendix~\ref{app:corpus}, Tables~\ref{tab:corpus-all} and~\ref{tab:corpus-paired}).

Furthermore, EIR's final answer call improves mean Answer F1 over the answer submitted during retrieval by 21.20 points using GPT-4.1 Mini and 6.88 using Haiku 4.5. This directly shows that a final model call can improve answers from already collected evidence (See Appendix~\ref{app:corpus-diagnostics}).

\begin{table}[!htb]
\centering\small
\begin{tabular*}{\textwidth}{@{\extracolsep{\fill}}lrrrr@{}}
\toprule
 & \multicolumn{2}{c}{GPT-4.1 Mini} & \multicolumn{2}{c}{Haiku 4.5} \\
\cmidrule(lr){2-3}\cmidrule(lr){4-5}
Method & Answer F1 & Calls/question & Answer F1 & Calls/question \\
\midrule
Direct & 48.30 & 6.19 & 57.07 & \textbf{6.33} \\
EIR & \textbf{61.67} & 7.61 & 61.35 & 12.03 \\
Agentic SSR & 49.14 & 70.29 & 52.02 & 88.04 \\
Reflexion, first trial & 51.51 & \textbf{6.17} & 50.38 & 8.30 \\
Reflexion, oracle stopping & 56.99 & 27.92 & \textbf{62.21} & 32.56 \\
\bottomrule
\end{tabular*}
\caption{\textbf{Extended HotpotQA evaluation requiring corpus search.} The same 1,000 questions are answered using each model, with evidence found among 5.23 million paragraphs. Bold marks the highest Answer F1 and fewest calls among displayed methods. Oracle-stopped Reflexion includes all preceding trials and reflections, while first-trial scores include just the first trial.}
\label{tab:corpus-main}
\end{table}

\begin{figure}[!htb]
\centering
\includegraphics[width=\textwidth]{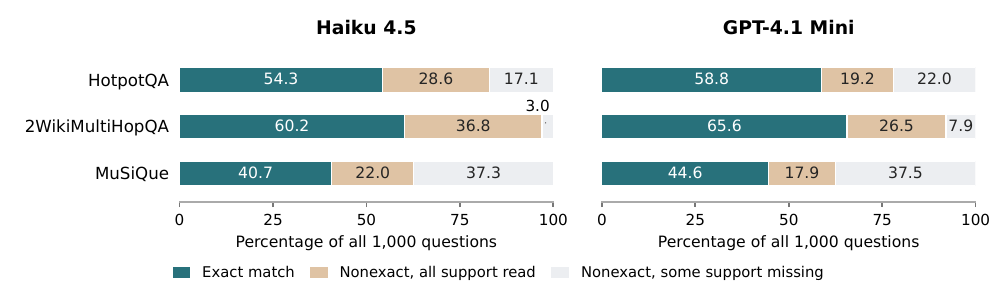}
\caption{\textbf{EIR answer outcomes across benchmarks and models.} Each bar represents 1,000 questions. Teal shows answers that exactly match the reference after normalization. The remaining answers are divided into those for which EIR read every annotated supporting paragraph (tan) and those missing at least one supporting paragraph (gray). Each percentage uses all 1,000 questions as its denominator, and the three portions sum to 100\%. The teal portion includes all exact matches without distinguishing their evidence coverage. Counts appear in Appendix~\ref{app:failure-analysis}, Table~\ref{tab:failure-phase-a}, and Appendix~\ref{app:confirmatory-results}, Table~\ref{tab:secondary}.}
\label{fig:evidence-outcomes}
\end{figure}

\paragraph{Evidence Availability and Remaining Errors.}
\textcolor{black}{Figure~\ref{fig:evidence-outcomes} shows the percentage of the time that EIR read all of the evidence necessary to answer each question. EIR did not read all of the necessary evidence in 17.1\%, 3.0\%, and 37.3\% of Haiku 4.5 answers on HotpotQA, 2WikiMultiHopQA, and MuSiQue, respectively, and in 22.0\%, 7.9\%, and 37.5\% of GPT-4.1 Mini answers. However, incorrect answers also occurred when EIR had read all of the necessary evidence: 28.6\%, 36.8\%, and 22.0\% for Haiku 4.5, and 19.2\%, 26.5\%, and 17.9\% for GPT-4.1 Mini, in the same benchmark order. Thus, EIR sometimes lacks information needed to answer the question, but it retrieves the evidence needed to answer the question most of the time.
} 

Adding missing supporting paragraphs to those incomplete cases and repeating the final answer call produces exact matches in 26.84\%, compared with 1.36\% when only the originally retrieved paragraphs are provided. A separate ablation lets the model choose up to two additional paragraphs, raising overall EM by 2.13 points but using about 37\% more calls and reversing some exact answers. The remaining challenge is to identify when further retrieval will help enough to justify its computation (See Appendix~\ref{app:failure-analysis}).

\section{Limitations}

\paragraph{Task Scope and Evidence Dependence.}
\textcolor{black}{We evaluated two language models, three multi-hop question-answering benchmarks with supplied questions and paragraphs, and one additional Wikipedia search setting. The recovery analyses also examine how additional evidence affects answers (See Appendix~\ref{app:failure-analysis}). We report computation for each method's retrieval and stopping procedure, and we evaluate each model and benchmark once. While we believe our evaluations have demonstrated convincingly the efficacy of EIR, the evaluation scope could be expanded to more language models and benchmarks.}

\paragraph{Attribution and Interfaces.}
\textcolor{black}{Direct, EIR, and the requirement and conclusion list baseline use different procedures for selecting evidence and producing answers. The reported comparisons therefore characterize complete methods, while the ablations provide targeted evidence about the role of the retrieval lists. The Wikipedia corpus search evaluation uses each method’s prescribed format for retrieval actions and supporting-sentence citations, and the reported scores include every trajectory (See Appendix~\ref{app:corpus-diagnostics}). A more uniform procedure could potentially be applied to all the compared methods. In spite of this, we believe that our experiments have been fairly executed to show the superiority of EIR.}

\section{Conclusion}

All in all, repeating intermediate reasoning steps or retrieval trajectories can be unnecessary when an agent has already collected the evidence needed to answer. EIR builds on this premise by using requirements and conclusions to guide one evidence retrieval trajectory, then generating the final answer from the accumulated facts without carrying forward the earlier intermediate reasoning or answers. Across HotpotQA, 2WikiMultiHopQA, and MuSiQue using both models, EIR significantly improves Answer F1 over Agentic SSR in all six comparisons and oracle-stopped Reflexion in five of six, while requiring substantially fewer calls than either method. The full-corpus evaluation further demonstrates accuracy and computation advantages over Agentic SSR using both models and oracle-stopped Reflexion using GPT-4.1 Mini. These findings support separating evidence acquisition from final inference, while the missing-evidence analyses motivate improving how the agent determines what information to retrieve. Future work should therefore strengthen both its ability to identify the facts and relationships required by a question and its ability to judge whether the retrieved evidence supports those requirements before ending retrieval. More reliable judgments could help the agent pursue overlooked information while avoiding additional retrieval once the necessary evidence is available.
\clearpage
\paragraph{Reproducibility.}
Appendices~\ref{app:conditions} and~\ref{app:repro} specify the conditions and implementation details, and Appendices~\ref{app:fullpool} and~\ref{app:corpus} document the additional evaluations and their fixed protocols.

\bibliographystyle{iclr2027_conference}

\paragraph*{Ethics Statement.}
This work evaluates language-model agents using established question-answering benchmarks and Wikipedia-derived text and does not involve the recruitment of human participants or the collection of new personal data. EIR is intended as a method for studying evidence acquisition and inference in multi-hop question answering, but its outputs can still be incorrect when relevant evidence is missing, misleading, or incorrectly interpreted. The method therefore should not be understood as establishing the reliability required for high-stakes decision-making or other settings in which unsupported answers could cause harm. In addition, EIR relies on pretrained language models and existing text corpora and may consequently inherit factual errors, representational biases, and other limitations present in those models and data. The experiments reported here evaluate answer accuracy and computational behavior rather than attempting to measure or mitigate these broader biases.

\paragraph*{AI Use Statement.}
In this work, the authors used generative AI tools, including ChatGPT and Claude Code, to assist with refining research hypotheses and experimental design, implementing and testing methods, orchestrating evaluation runs, developing statistical-analysis and result-auditing scripts, interpreting experimental results, and drafting and editing manuscript prose, tables, figures, and \LaTeX{}. All AI-assisted code and analyses were reviewed and validated through automated tests, record-integrity checks, and recomputation from saved per-question outcomes, and citations, numerical results, methodological descriptions, and factual claims were checked against the referenced papers, implementation, and recorded evaluation artifacts. Generative AI was not used to create benchmark questions, reference answers, supporting-fact annotations, or evaluation labels. We have reviewed all AI-assisted work and take responsibility for the final content of this work, including text, claims, and artifacts produced with its aid.

\clearpage
\clearpage
\appendix

\noindent
\begin{tabular*}{\textwidth}{@{\extracolsep{\fill}}lr@{}}
\toprule
Appendix contents & Page \\
\midrule
\hyperref[app:trace-comparison]{\ref*{app:trace-comparison}. Recorded SSR and Reflexion traces} & \pageref{app:trace-comparison} \\
\hyperref[app:protocol]{\ref*{app:protocol}. Method and evaluation details} & \pageref{app:protocol} \\
\hyperref[app:confirmatory-results]{\ref*{app:confirmatory-results}. Additional results of the main evaluation} & \pageref{app:confirmatory-results} \\
\hyperref[app:comparison]{\ref*{app:comparison}. Matched SSR and Reflexion comparisons} & \pageref{app:comparison} \\
\hyperref[app:mechanism-ablations]{\ref*{app:mechanism-ablations}. Retrieval lists and dependency repair} & \pageref{app:mechanism-ablations} \\
\hyperref[app:failure-analysis]{\ref*{app:failure-analysis}. Missing evidence and additional retrieval} & \pageref{app:failure-analysis} \\
\hyperref[app:fullpool]{\ref*{app:fullpool}. Full Pool evaluation} & \pageref{app:fullpool} \\
\hyperref[app:corpus]{\ref*{app:corpus}. Full-corpus evaluation and error analysis} & \pageref{app:corpus} \\
\bottomrule
\end{tabular*}
\par\medskip

\section{Recorded SSR and Reflexion Traces}
\label{app:trace-comparison}

\begin{figure}[!htbp]
\centering
\includegraphics[width=\textwidth,height=0.78\textheight,keepaspectratio]{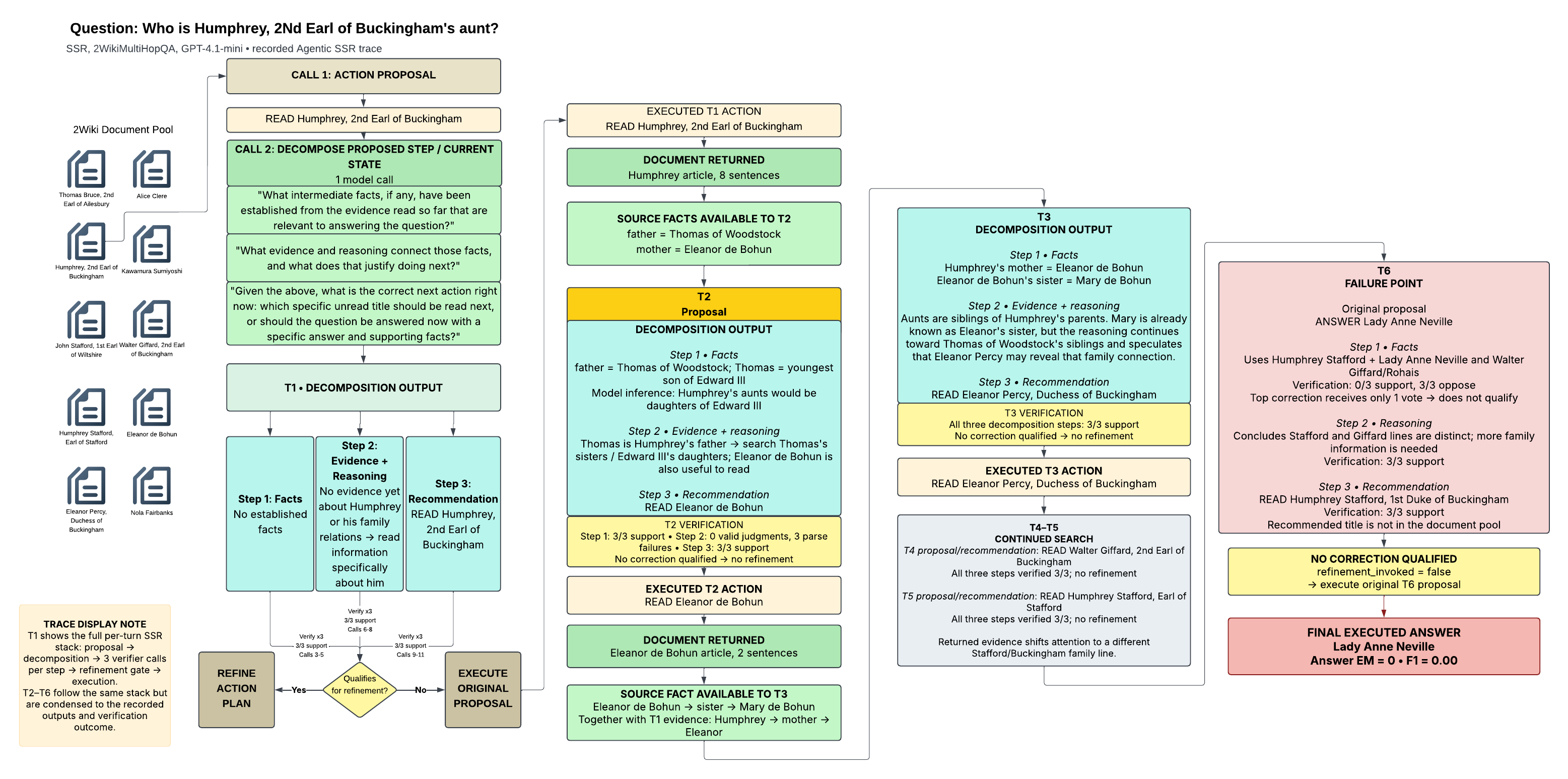}
\caption{\textbf{Agentic SSR on the Humphrey question.} The first two reads establish the relationships needed to identify his aunt. Subsequent actions pursue his father's relatives and unrelated family branches. On the final turn, all three judgments reject the factual reasoning step, but their proposed corrections do not meet the agreement rule for refinement, and the agent submits Lady Anne Neville. The diagram shows all verification calls for the first turn and condenses later turns.}
\label{fig:ssr-recorded-trace}
\end{figure}

\begin{figure}[!htbp]
\centering
\includegraphics[width=\textwidth,height=0.78\textheight,keepaspectratio]{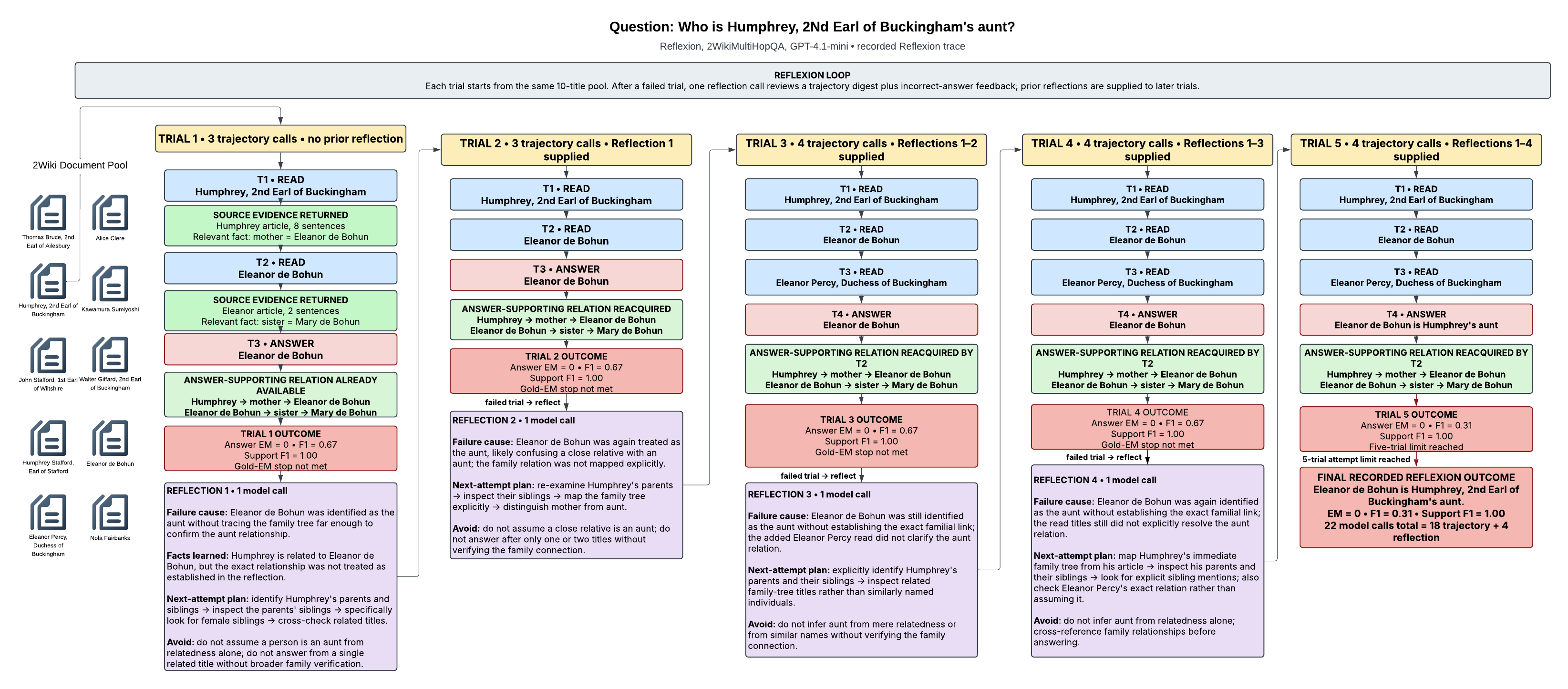}
\caption{\textbf{Reflexion on the Humphrey question.} All five trials read the supporting relationships but identify Humphrey's mother, Eleanor de Bohun, as his aunt. The intervening reflections call for further verification of his family relationships. None of the answers matches the reference, so all five trials run, using 18 action calls and four reflection calls.}
\label{fig:reflexion-recorded-trace}
\end{figure}

Figures~\ref{fig:ssr-recorded-trace} and~\ref{fig:reflexion-recorded-trace} show recorded runs using GPT-4.1 Mini on the 2WikiMultiHopQA question illustrated for EIR in Figure~\ref{fig:eir-recorded-trace}: \textit{Who is Humphrey, 2nd Earl of Buckingham's aunt?} The retrieved paragraphs identify Eleanor de Bohun as Humphrey's mother and Mary de Bohun as Eleanor's sister. Agentic SSR and Reflexion both read these relationships but return incorrect answers.

The EIR trace in Figure~\ref{fig:eir-recorded-trace} comes from a separate run recorded for illustration, outside the scored evaluation. It uses six paragraph reads, one submission call, and one final answer call, for eight calls in total. Its source evidence and final lists are condensed for display. The main evaluation's configured budget of $H=6$ likewise permits six reads followed by submission, rather than limiting the entire run to six actions.

\section{Method and Evaluation Details}
\label{app:protocol}

\subsection{Evidence Records, Questions, and Baselines}
\label{app:rationale}
\label{app:impl}
\label{app:conditions}

Each requirement and conclusion has an identifier and status, and each conclusion cites sentences already read. EIR checks proposed changes together and leaves both lists unchanged if any part is invalid. It checks that citations identify existing sentences, not whether those sentences justify the claim. Source text is copied into separate records containing the paragraph title, sentence number, and retrieval turn, and the model cannot rewrite it. Default EIR has no links between conclusions. Appendix~\ref{app:repair} tests adding such links.

Agents select from unread titles. If a requested title does not match, the environment reads the first unread title in the provided order. EIR permits six reads and, if needed, another decision to submit. Its separate final answer call then receives only the question and collected source sentences.

The main evaluation uses the development questions and paragraphs released with each benchmark: ten paragraphs per question for HotpotQA and 2WikiMultiHopQA, and twenty for answerable MuSiQue. Their text, sentence boundaries, and order are preserved. Agents initially see only titles. Answering and reflection prompts never include reference answers or labels identifying the evidence needed.

Every method and both models use the same 1,000 questions per benchmark, selected from questions unused in earlier experiments. Twenty separate questions per benchmark test the setup and are excluded from scoring. In the tables, 2Wiki means 2WikiMultiHopQA, CI means confidence interval, and pp means percentage points.

\subsection{Model and Analysis Settings}
\label{app:repro}

Haiku 4.5 uses a 2,048-token output limit for Direct and EIR, overriding the configured limits in Table~\ref{tab:repro}. For GPT-4.1 Mini, the table reports configured limits because complete historical requests were not retained.

\begingroup\footnotesize
\setlength{\LTcapwidth}{\textwidth}
\begin{longtable}{p{0.33\textwidth}p{0.60\textwidth}}
\caption{Question selection, model, statistical, and cost settings for the main evaluation. Output limits count generated tokens, and dollar costs are reconstructed at the fixed prices below rather than taken from invoices. Full Pool and full-corpus settings appear in Appendices~\ref{app:fullpool} and~\ref{app:corpus}.}
\label{tab:repro}\label{tab:reproB}\\
\toprule
Setting & Value \\
\midrule\endfirsthead
\toprule Setting & Value \\
\midrule\endhead
\bottomrule\endfoot\endlastfoot
\multicolumn{2}{l}{\emph{Questions and models}} \\*
Model snapshots & \texttt{claude-haiku-4-5-20251001} and \texttt{gpt-4.1-mini-2025-04-14} \\
HotpotQA & Distractor development split: 7,404 eligible questions, 6,068 unused earlier, 1,000 selected \\
2WikiMultiHopQA & Development split: 12,576 eligible questions, 11,506 unused earlier, 1,000 selected \\
MuSiQue & Answerable development questions: 2,417 eligible, 1,347 unused earlier, 1,000 selected \\
Selection & Questions drawn in sorted order from those unused in earlier evaluations, using fixed question lists \\
Calibration & 20 separate questions per benchmark, excluded from scoring \\
\midrule
\multicolumn{2}{l}{\emph{Generation settings}} \\*
EIR retrieval budget & $H=6$: at most six reads, followed by submission if necessary. The final answer call is counted separately. \\
Direct action calls & Temperature 0.0, configured output limit 320 tokens \\
EIR action calls & Temperature 0.0, configured output limit 900 tokens \\
EIR final answer call & Temperature 0.0, configured output limit 500 tokens \\
Haiku 4.5 override & Effective output limit of 2,048 tokens for the Direct and EIR calls above \\
Dependency-repair baselines & Temperature 0.0. Configured limits: 900 action tokens, 400 repair tokens, 400 requirement-refresh tokens, and 500 answer tokens. At most ten model calls, two repairs, and two additional answer calls per episode. \\
Retry for an invalid format & None for Direct or default EIR. One permitted in the dependency-repair baseline. \\
Request timeout and retries & 60-second timeout, no automatic client retry in the original evaluation \\
Model sampling seed & None, because the client does not support one \\
\midrule
\multicolumn{2}{l}{\emph{Statistics and cost}} \\*
Answer F1 comparisons & 10,000 paired bootstrap resamples of questions, with percentile 95\% confidence intervals \\
Bootstrap seeds & 20260804 for the original evaluation and the SSR and Reflexion comparisons, 20260806 for the list-feedback and repair-application analyses \\
Exact Match comparisons & Exact two-sided McNemar tests, using questions on which only one method matches the reference answer \\
Multiple comparisons & Holm adjustment within the groups of comparisons named in each table caption \\
Haiku 4.5 token prices & \$1.00 per million input tokens and \$5.00 per million output tokens \\
GPT-4.1 Mini token prices & \$0.40 per million input tokens and \$1.60 per million output tokens \\
\bottomrule
\end{longtable}
\endgroup

\subsection{Prompt Inputs and Answer Generation}
\label{app:prompt-inputs}

EIR's action prompt includes the question, the previous action's returned text or feedback, both lists, and available titles or search results. One call proposes the next action and list updates. Direct omits the lists. Agentic SSR adds verification and possible refinement calls, while Reflexion carries written feedback into later trials (Appendix~\ref{app:comparison-adaptation}).

EIR's final prompt contains only the question and collected sentences, labeled by paragraph and sentence position for citation. It excludes earlier answers, lists, and reasoning. Full Pool uses the same answering instructions and evidence format with every provided paragraph, including distractors (Appendix~\ref{app:fullpool}).

During corpus search, Direct's action prompt also asks what the proposed action should establish, its expected result, and how progress will be judged. It includes the two most recently read paragraphs. EIR uses its lists to guide searches and reads, then answers from all collected text.

\section{Additional Results of the Main Evaluation}
\label{app:confirmatory-results}

\subsection{Answer Accuracy and Supporting Citations}
\label{app:em}
\label{app:support}

Table~\ref{tab:secondary} adds Exact Match and Support F1 to the main Answer F1 results. All scores include 1,000 questions per benchmark and model. MuSiQue Exact Match is an additional measure used in this study.

\begin{table}[H]
\centering
\footnotesize
\begin{tabular*}{\textwidth}{@{\extracolsep{\fill}}llrrrlr@{}}
\toprule
& & \multicolumn{2}{c}{Answer EM} & \multicolumn{3}{c}{Support F1} \\
\cmidrule(lr){3-4}\cmidrule(lr){5-7}
Model & Benchmark & Direct & \eir{} & Direct & \eir{} & 95\% CI (diff.) \\
\midrule
\multirow{3}{*}{Haiku 4.5}
& HotpotQA & 45.0 & \textbf{54.3} & 68.80 & \textbf{74.50} & [$+4.13$, $+7.18$] \\
& 2Wiki & 33.0 & \textbf{60.2} & 66.15 & \textbf{85.69} & [$+17.18$, $+21.83$] \\
& MuSiQue & 31.5 & \textbf{40.7} & 48.83 & \textbf{67.64} & [$+16.35$, $+21.19$] \\
\midrule
\multirow{3}{*}{GPT-4.1 Mini}
& HotpotQA & 34.5 & \textbf{58.8} & 65.45 & \textbf{68.67} & [$+1.67$, $+4.81$] \\
& 2Wiki & 21.7 & \textbf{65.6} & 62.93 & \textbf{72.67} & [$+7.71$, $+11.83$] \\
& MuSiQue & 19.5 & \textbf{44.6} & 56.75 & \textbf{62.50} & [$+3.86$, $+7.61$] \\
\bottomrule
\end{tabular*}
\caption{\textbf{Exact Match and Support F1 for Direct and EIR.} Each row includes all 1,000 questions, and bold marks the higher score in each pair. Intervals are paired bootstrap 95\% intervals for EIR minus Direct Support F1, and each excludes zero.}
\label{tab:secondary}\label{tab:support-paired}
\end{table}

\subsection{Outputs That Could Not Be Parsed}
\label{app:parse}

Outputs that could not be read in the required answer format account for 2.8\%--20.0\% of Direct responses using Haiku 4.5 and 1.5\%--5.3\% using GPT-4.1 Mini. The corresponding EIR ranges are 0.0\%--0.1\% and 1.0\%--1.4\%. All questions remain in the main scores.

Keeping only questions with valid outputs from both methods reduces the MuSiQue Answer F1 gain using Haiku 4.5 from 11.16 to 3.76 points over 800 questions (95\% CI $[1.30,6.17]$, $p=0.0018$). The other five gains range from 6.76 to 30.65 points. All six intervals exclude zero, and the Exact Match conclusions remain unchanged. These results use smaller question sets than the main evaluation.

\section{Matched SSR and Reflexion Comparisons}
\label{app:comparison}

Agentic SSR and Reflexion answer the same 1,000 questions per benchmark and model as EIR. These comparisons were added after the original evaluation was designed.

\subsection{Adaptation to Action Selection}
\label{app:comparison-adaptation}

Agentic SSR first proposes an action. Another call divides its reasoning into three parts: established facts, how those facts justify the action, and the action to execute. Three verifier calls assess each part at temperature 0.7 \citep{shi2025ssr}. Refinement requires at least two rejections, a majority of valid judgments rejecting the part, and a majority of those rejections agreeing on a correction. At most one call refines the qualifying part with the lowest confidence, breaking ties in favor of the later part. If refinement gives no valid action, an agreed correction to the action part is used if available. Otherwise, the original action is retained. Invalid judgments do not vote. Unlike published SSR-Lin, which refines a completed response three times, this adaptation permits one refinement per action.

Reflexion repeats retrieval with written feedback from earlier failed trials \citep{shinn2023reflexion}. Its first trial measures performance before any feedback or retry. An external controller stops oracle-stopped Reflexion at the first exact match to the reference answer, or returns trial five if none matches. The model never receives the reference answer, and the controller does not choose the highest Answer F1. Our five-trial limit and provided paragraphs differ from the published HotpotQA implementation's Wikipedia API, memory of up to three attempts, and limit of three consecutive failed attempts. Additional verification, refinement, and reflection calls count toward each method's computation.

\subsection{Answer F1}

Table~\ref{tab:comparisonf1} subtracts each baseline's Answer F1 from EIR's on the same questions. Positive differences favor EIR.

\begingroup
\footnotesize
\setlength{\tabcolsep}{4pt}
\setlength{\LTpre}{6pt}
\setlength{\LTpost}{6pt}
\setlength{\LTcapwidth}{\textwidth}
\begin{longtable}{llrlrrc}
\caption{Answer F1 differences, EIR minus each baseline, in percentage points. Each row compares both methods on the same 1,000 questions, with paired bootstrap 95\% intervals. Holm adjustment covers the six rows for each baseline, and values below the reporting resolution appear as $<10^{-4}$.}
\label{tab:comparisonf1}\\
\toprule
Model & Benchmark & $\Delta$F1 (pp) & 95\% CI (pp) & raw $p$ & Holm $p$ & sig.\ \\
\midrule
\endfirsthead
\toprule
Model & Benchmark & $\Delta$F1 (pp) & 95\% CI (pp) & raw $p$ & Holm $p$ & sig.\ \\
\midrule
\endhead
\bottomrule
\endfoot
\endlastfoot
\multicolumn{7}{l}{\emph{\eir{} vs.\ Agentic SSR}} \\*
Haiku 4.5 & HotpotQA & $+12.59$ & [$+10.46$, $+14.73$] & $<10^{-4}$ & $<10^{-4}$ & yes \\
Haiku 4.5 & 2Wiki & $+29.12$ & [$+26.34$, $+31.82$] & $<10^{-4}$ & $<10^{-4}$ & yes \\
Haiku 4.5 & MuSiQue & $+10.56$ & [$+8.22$, $+12.86$] & $<10^{-4}$ & $<10^{-4}$ & yes \\
GPT-4.1 Mini & HotpotQA & $+14.54$ & [$+12.40$, $+16.70$] & $<10^{-4}$ & $<10^{-4}$ & yes \\
GPT-4.1 Mini & 2Wiki & $+26.66$ & [$+24.08$, $+29.32$] & $<10^{-4}$ & $<10^{-4}$ & yes \\
GPT-4.1 Mini & MuSiQue & $+22.72$ & [$+20.12$, $+25.28$] & $<10^{-4}$ & $<10^{-4}$ & yes \\
\midrule
\multicolumn{7}{l}{\emph{\eir{} vs.\ Reflexion trial 1 (first attempt, no reference answer)}} \\*
Haiku 4.5 & HotpotQA & $+11.14$ & [$+9.08$, $+13.20$] & $<10^{-4}$ & $<10^{-4}$ & yes \\
Haiku 4.5 & 2Wiki & $+21.18$ & [$+18.48$, $+23.84$] & $<10^{-4}$ & $<10^{-4}$ & yes \\
Haiku 4.5 & MuSiQue & $+6.04$ & [$+3.69$, $+8.34$] & $<10^{-4}$ & $<10^{-4}$ & yes \\
GPT-4.1 Mini & HotpotQA & $+12.58$ & [$+10.42$, $+14.76$] & $<10^{-4}$ & $<10^{-4}$ & yes \\
GPT-4.1 Mini & 2Wiki & $+18.96$ & [$+16.65$, $+21.30$] & $<10^{-4}$ & $<10^{-4}$ & yes \\
GPT-4.1 Mini & MuSiQue & $+19.89$ & [$+17.25$, $+22.48$] & $<10^{-4}$ & $<10^{-4}$ & yes \\
\midrule
\multicolumn{7}{l}{\emph{\eir{} vs.\ oracle-stopped Reflexion}} \\*
Haiku 4.5 & HotpotQA & $+4.98$ & [$+2.86$, $+7.14$] & $<10^{-4}$ & $<10^{-4}$ & yes \\
Haiku 4.5 & 2Wiki & $+4.26$ & [$+1.88$, $+6.68$] & $2.0\times10^{-4}$ & $4.0\times10^{-4}$ & yes \\
Haiku 4.5 & MuSiQue & $+1.13$ & [$-1.36$, $+3.62$] & $0.372$ & $0.372$ & \textbf{no} \\
GPT-4.1 Mini & HotpotQA & $+8.92$ & [$+6.61$, $+11.25$] & $<10^{-4}$ & $<10^{-4}$ & yes \\
GPT-4.1 Mini & 2Wiki & $+15.92$ & [$+13.47$, $+18.43$] & $<10^{-4}$ & $<10^{-4}$ & yes \\
GPT-4.1 Mini & MuSiQue & $+14.75$ & [$+11.97$, $+17.49$] & $<10^{-4}$ & $<10^{-4}$ & yes \\*
\bottomrule
\end{longtable}
\endgroup

\subsection{Exact Match}

Table~\ref{tab:comparisonem} counts questions on which only one method produces an exact match. On MuSiQue using Haiku 4.5, oracle-stopped Reflexion has higher Exact Match even though EIR's Answer F1 point estimate is higher. The metrics differ because Answer F1 gives partial credit for word overlap, whereas Exact Match does not.

\begingroup
\footnotesize
\setlength{\tabcolsep}{4pt}
\setlength{\LTpre}{6pt}
\setlength{\LTpost}{6pt}
\setlength{\LTcapwidth}{\textwidth}
\begin{longtable}{llrrr}
\caption{Exact Match differences, EIR minus each baseline, in percentage points. Each row compares both methods on the same 1,000 questions, and in $b/c$, $b$ counts questions answered exactly only by the baseline and $c$ those answered exactly only by EIR. $p$-values are unadjusted exact two-sided McNemar tests.}
\label{tab:comparisonem}\\
\toprule
Model & Benchmark & $\Delta$EM (pp) & Only one exact ($b$/$c$) & McNemar $p$ \\
\midrule
\endfirsthead
\toprule
Model & Benchmark & $\Delta$EM (pp) & Only one exact ($b$/$c$) & McNemar $p$ \\
\midrule
\endhead
\bottomrule
\endfoot
\endlastfoot
\multicolumn{5}{l}{\emph{\eir{} vs.\ Agentic SSR}} \\*
Haiku 4.5 & HotpotQA & $+12.90$ & 28/157 & $<10^{-4}$ \\
Haiku 4.5 & 2Wiki & $+34.90$ & 15/364 & $<10^{-4}$ \\
Haiku 4.5 & MuSiQue & $+9.60$ & 38/134 & $<10^{-4}$ \\
GPT-4.1 Mini & HotpotQA & $+21.40$ & 28/242 & $<10^{-4}$ \\
GPT-4.1 Mini & 2Wiki & $+36.70$ & 19/386 & $<10^{-4}$ \\
GPT-4.1 Mini & MuSiQue & $+23.60$ & 23/259 & $<10^{-4}$ \\
\midrule
\multicolumn{5}{l}{\emph{\eir{} vs.\ Reflexion trial 1}} \\*
Haiku 4.5 & HotpotQA & $+10.70$ & 35/142 & $<10^{-4}$ \\
Haiku 4.5 & 2Wiki & $+25.50$ & 30/285 & $<10^{-4}$ \\
Haiku 4.5 & MuSiQue & $+4.90$ & 56/105 & $1.39\times10^{-4}$ \\
GPT-4.1 Mini & HotpotQA & $+18.80$ & 33/221 & $<10^{-4}$ \\
GPT-4.1 Mini & 2Wiki & $+28.80$ & 24/312 & $<10^{-4}$ \\
GPT-4.1 Mini & MuSiQue & $+21.70$ & 31/248 & $<10^{-4}$ \\
\midrule
\multicolumn{5}{l}{\emph{\eir{} vs.\ oracle-stopped Reflexion}} \\*
Haiku 4.5 & HotpotQA & $+3.20$ & 69/101 & $1.72\times10^{-2}$ \\
Haiku 4.5 & 2Wiki & $+8.30$ & 75/158 & $<10^{-4}$ \\
Haiku 4.5 & MuSiQue & $-3.90$ & 99/60 & $2.47\times10^{-3}$ \\
GPT-4.1 Mini & HotpotQA & $+12.70$ & 64/191 & $<10^{-4}$ \\
GPT-4.1 Mini & 2Wiki & $+22.70$ & 44/271 & $<10^{-4}$ \\
GPT-4.1 Mini & MuSiQue & $+12.70$ & 74/201 & $<10^{-4}$ \\*
\bottomrule
\end{longtable}
\endgroup

\subsection{Supporting Citations}

Against oracle-stopped Reflexion, EIR has significantly higher Support F1 on all three benchmarks using Haiku 4.5. Using GPT-4.1 Mini, the difference favors oracle-stopped Reflexion on all three benchmarks, significantly so only on 2WikiMultiHopQA (Table~\ref{tab:comparisonsupport}).

\begingroup
\footnotesize
\setlength{\tabcolsep}{4pt}
\setlength{\LTpre}{6pt}
\setlength{\LTpost}{6pt}
\setlength{\LTcapwidth}{\textwidth}
\begin{longtable}{llrrrrrl}
\caption{Support F1 on a 0--100 scale. Each row includes all 1,000 questions, where SSR is Agentic SSR, Refl.\ t1 is Reflexion's first trial, and Refl.\ oracle is oracle-stopped Reflexion. The last two columns give EIR-minus-oracle differences with Holm adjustment across those six comparisons.}
\label{tab:comparisonsupport}\\
\toprule
Model & Benchmark & \eir{} & SSR & Refl.\ t1 & Refl.\ oracle & $\Delta$(\eir{}$-$oracle) & Holm $p$, sig.\ \\
\midrule
\endfirsthead
\toprule
Model & Benchmark & \eir{} & SSR & Refl.\ t1 & Refl.\ oracle & $\Delta$(\eir{}$-$oracle) & Holm $p$, sig.\ \\
\midrule
\endhead
\bottomrule
\endfoot
\endlastfoot
\multirow{3}{*}{Haiku 4.5}
& HotpotQA & 74.50 & 68.73 & 71.00 & 70.34 & $+4.16$ & $<10^{-4}$, yes \\*
& 2Wiki & 85.69 & 67.30 & 77.17 & 79.26 & $+6.44$ & $<10^{-4}$, yes \\*
& MuSiQue & 67.64 & 51.22 & 61.23 & 58.44 & $+9.21$ & $<10^{-4}$, yes \\
\midrule
\multirow{3}{*}{GPT-4.1 Mini}
& HotpotQA & 68.67 & 64.07 & 67.08 & 69.36 & $-0.68$ & $0.777$, no \\*
& 2Wiki & 72.67 & 62.16 & 71.52 & 74.97 & $-2.31$ & $0.009$, \textbf{yes} \\*
& MuSiQue & 62.50 & 53.28 & 59.91 & 63.38 & $-0.88$ & $0.777$, no \\*
\bottomrule
\end{longtable}
\endgroup

\subsection{Computation}

Table~\ref{tab:computation} averages over the three benchmarks and two models equally. Calls include all steps producing an answer, including previous Reflexion trials and reflections, but exclude provider retries. Some GPT-4.1 Mini reflection tokens lack an input/output breakdown. The cost range either omits those tokens or prices them all as output. Haiku 4.5 has complete token counts.

\begin{table}[H]
\centering
\footnotesize
\setlength{\tabcolsep}{3pt}
\begin{tabular*}{\textwidth}{@{\extracolsep{\fill}}lrrrrr@{}}
\toprule
Method & Answer F1 & Calls/q & Calls/EIR & \$/q & Cost/EIR \\
\midrule
EIR & \textbf{66.89} & 4.85 & 1.00$\times$ & 0.00738 & 1.00$\times$ \\
Agentic SSR & 47.52 & 35.29 & 7.28$\times$ & 0.02681 & 3.63$\times$ \\
Reflexion (oracle) & 58.56 & 12.41 & 2.56$\times$ & 0.01149--0.01274 & 1.56--1.73$\times$ \\
Direct & 47.93 & \textbf{3.24} & \textbf{0.67}$\times$ & \textbf{0.00311} & \textbf{0.42}$\times$ \\
\bottomrule
\end{tabular*}
\caption{\textbf{Average accuracy and computation in the main evaluation.} Rows are methods averaged over the three benchmarks and two models, with ratios relative to the EIR mean and dollar costs at the fixed prices in Table~\ref{tab:reproB}. Bold marks the highest Answer F1, fewest calls, and lowest cost, and the Reflexion cost range reflects missing token breakdowns rather than statistical uncertainty.}
\label{tab:computation}
\end{table}

\section{Retrieval Lists and Dependency Repair}
\label{app:mechanism-ablations}
\label{app:comparisons}

This appendix examines EIR's read budget, the contribution of the requirement and conclusion lists, and the effect of revising dependent conclusions.

\subsection{Turn Budget Sensitivity Analysis}
\label{app:turn-budget-analysis}
\begin{figure}[H]
\centering
\includegraphics[width=\textwidth]{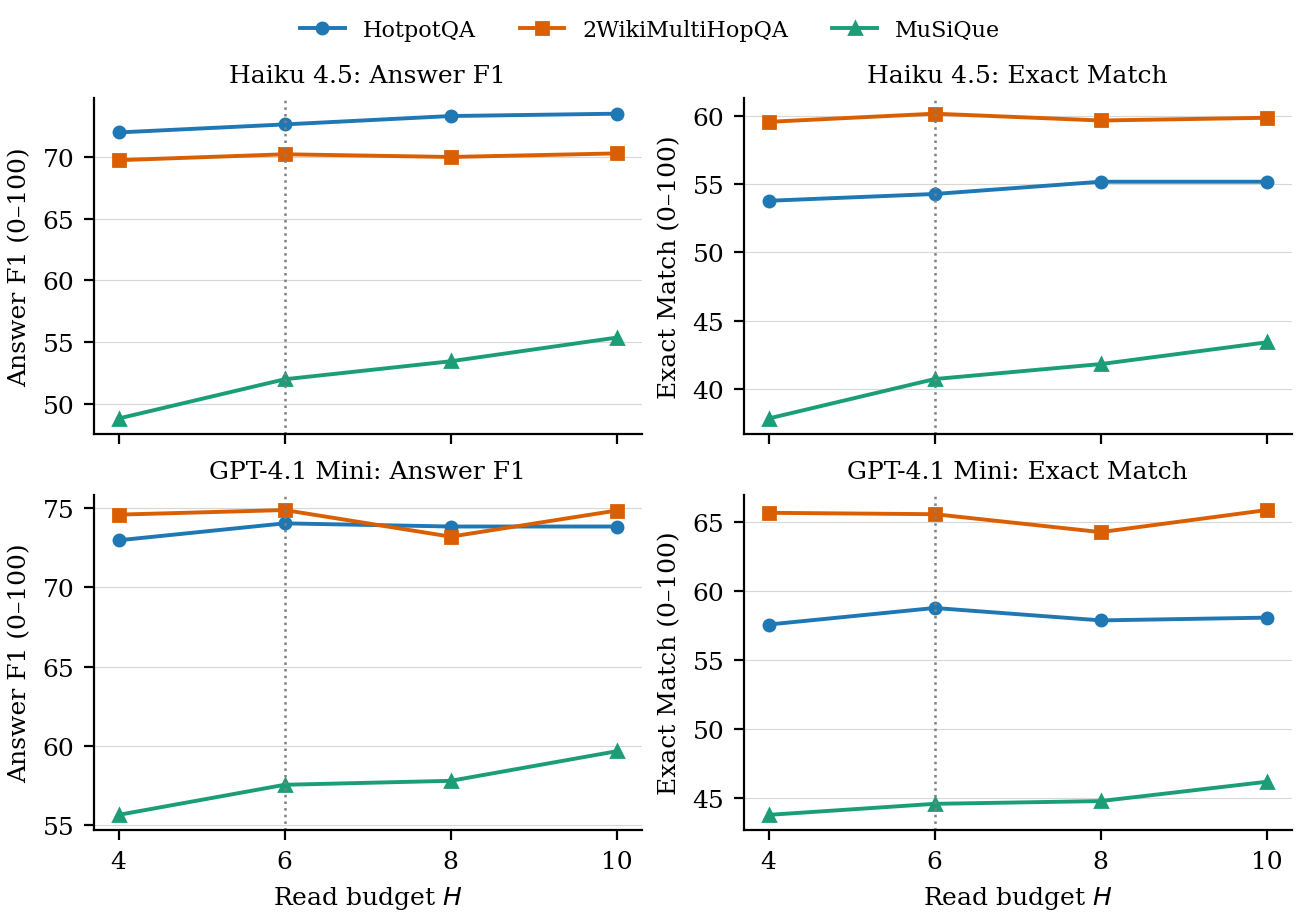}
\caption{\textbf{EIR accuracy across read budgets.} Each point averages the same 1,000 questions per benchmark and model. The dotted line marks the budget used in the main evaluation, $H=6$.}
\label{fig:turn-budget}
\end{figure}

The main evaluation limits EIR to $H=6$ paragraph reads, a value fixed before the evaluation began. Submission can follow as one additional action, and the final answer call runs after it. To test how sensitive EIR is to the choice of $H$, we repeat the main evaluation using both models with $H$ set to 4, 6, 8, and 10, keeping the questions, prompts, and final answer call unchanged. Figure~\ref{fig:turn-budget} reports Answer F1 and Exact Match for each budget, Table~\ref{tab:turn-budget-calls} reports the mean number of model calls per question, including the final answer call, and Table~\ref{tab:turn-budget-paired} compares each budget with $H=6$.

\begin{table}[H]
\centering\small
\begin{tabular*}{\textwidth}{@{\extracolsep{\fill}}llrrrrr@{}}
\toprule
Model & Benchmark & $H=4$ & $H=6$ & $H=8$ & $H=10$ & $H=10$ vs.\ $H=6$ \\
\midrule
Haiku 4.5 & HotpotQA & 4.24 & 4.52 & 4.83 & 5.09 & $+12.6\%$ \\
 & 2WikiMultiHopQA & 4.74 & 5.11 & 5.42 & 5.78 & $+13.1\%$ \\
 & MuSiQue & 4.76 & 5.32 & 5.79 & 6.14 & $+15.4\%$ \\
\midrule
GPT-4.1 Mini & HotpotQA & 3.91 & 4.08 & 4.14 & 4.20 & $+3.1\%$ \\
 & 2WikiMultiHopQA & 4.54 & 4.86 & 5.04 & 5.27 & $+8.4\%$ \\
 & MuSiQue & 4.54 & 5.19 & 5.73 & 6.00 & $+15.6\%$ \\
\bottomrule
\end{tabular*}
\caption{\textbf{Mean model calls per question across read budgets.} Each value averages 1,000 questions and includes the final answer call. The last column gives the relative increase from $H=6$ to $H=10$.}
\label{tab:turn-budget-calls}
\end{table}

\begin{table}[H]
\centering\small
\begin{tabular*}{\textwidth}{@{\extracolsep{\fill}}llrrr@{}}
\toprule
Model & Benchmark & $H=4$ & $H=8$ & $H=10$ \\
\midrule
Haiku 4.5 & HotpotQA & $-0.65$ & $+0.68$ & $+0.87$ \\
 & 2WikiMultiHopQA & $-0.48$ & $-0.21$ & $+0.08$ \\
 & MuSiQue & $-3.18^{*}$ & $+1.46$ & $+3.37^{*}$ \\
\midrule
GPT-4.1 Mini & HotpotQA & $-1.07$ & $-0.20$ & $-0.20$ \\
 & 2WikiMultiHopQA & $-0.29$ & $-1.68$ & $-0.03$ \\
 & MuSiQue & $-1.90$ & $+0.25$ & $+2.11$ \\
\bottomrule
\end{tabular*}
\caption{\textbf{Answer F1 change relative to $H=6$.} Each value is the named budget minus $H=6$ on the same 1,000 questions, in percentage points, calculated before rounding. Differences use paired bootstrap tests with 10,000 resamples, and Holm adjustment covers all 18 comparisons. $^{*}$ marks significance after adjustment.}
\label{tab:turn-budget-paired}
\end{table}

On HotpotQA and 2WikiMultiHopQA, no budget changes Answer F1 significantly relative to $H=6$ using either model, and every difference stays within 1.7 points. Raising the budget beyond $H=6$ therefore adds calls on these two benchmarks without a detectable accuracy gain. MuSiQue behaves differently. Its questions can require up to four supporting paragraphs from a pool of twenty, and Answer F1 rises with the budget using both models. Using Haiku 4.5, $H=10$ improves Answer F1 by 3.37 points (95\% CI $[+1.90,+4.85]$) and $H=4$ lowers it by 3.18 points ($[-4.71,-1.64]$), both significant after adjustment, at the cost of 15.4\% more calls for $H=10$. Using GPT-4.1 Mini, $H=10$ improves Answer F1 by 2.11 points ($[+0.26,+3.93]$), but the difference is not significant after adjustment.

This sensitivity check supports our pre-existing choice of $H=6$. It reaches accuracy statistically indistinguishable from larger budgets on HotpotQA and 2WikiMultiHopQA, while larger budgets mainly benefit questions that require more supporting paragraphs, as in MuSiQue, at a measurable increase in calls. The main evaluation's MuSiQue scores may therefore understate EIR's accuracy when more reads are allowed.
\subsection{List-Feedback Control}
\label{app:listfeedback}

The list-feedback-disabled control tests whether including \eir{}'s requirement and conclusion lists in later action-selection prompts improves accuracy. Including them increases Answer F1 on 2WikiMultiHopQA and MuSiQue with both models, while neither HotpotQA comparison shows a positive effect (Table~\ref{tab:listfeedback}). Gold-document recall, the fraction of the benchmark's annotated supporting paragraphs that the agent reads, does not consistently rise when the lists are included, so these gains are not explained by reading more of those paragraphs.

For \eir{} versus the control, gold-document recall is 83.75\% versus 91.00\%, 97.75\% versus 97.82\%, and 75.95\% versus 77.90\% using Haiku 4.5 on HotpotQA, 2WikiMultiHopQA, and MuSiQue, and 77.90\% versus 81.40\%, 93.80\% versus 93.23\%, and 72.33\% versus 73.47\% using GPT-4.1 Mini. The control reads more paragraphs per question in every comparison, averaging 5.03 versus 2.52, 5.72 versus 3.11, and 5.54 versus 3.32 using Haiku 4.5 on HotpotQA, 2WikiMultiHopQA, and MuSiQue, and 3.37 versus 2.08, 4.43 versus 2.86, and 3.99 versus 3.19 using GPT-4.1 Mini. Its recall is therefore higher on HotpotQA by 7.25 and 3.50 points and on MuSiQue by 1.95 and 1.13 points, while EIR and the control differ by less than one point on 2WikiMultiHopQA. 

\begingroup
\footnotesize
\setlength{\tabcolsep}{5pt}
\setlength{\LTpre}{6pt}
\setlength{\LTpost}{6pt}
\setlength{\LTcapwidth}{\textwidth}

\begin{table}[H]
\centering\footnotesize
\setlength{\tabcolsep}{4pt}
\begin{tabular*}{\textwidth}{@{\extracolsep{\fill}}llrrrlr@{}}
\toprule
Model & Benchmark & \shortstack[r]{Corrected\\control} & EIR & Difference & 95\% CI & Holm $p$ \\
\midrule
Haiku 4.5 & HotpotQA & 73.91 & 72.64 & $-1.27$ & [$-2.71$, $+0.12$] & 0.2256 \\
 & 2Wiki & 69.41 & 70.22 & $+0.80$ & [$-0.48$, $+2.13$] & 0.4120 \\
 & MuSiQue & 47.15 & 51.98 & $+4.83^{*}$ & [$+2.79$, $+6.86$] & $<0.001$ \\
\midrule
GPT-4.1 Mini & HotpotQA & 75.49 & 74.05 & $-1.44$ & [$-2.83$, $-0.06$] & 0.1624 \\
 & 2Wiki & 72.45 & 74.89 & $+2.45^{*}$ & [$+1.08$, $+3.89$] & 0.0060 \\
 & MuSiQue & 56.25 & 57.54 & $+1.30$ & [$-0.74$, $+3.28$] & 0.4120 \\
\bottomrule
\end{tabular*}
\caption{\textbf{Answer F1 with and without the lists in later retrieval prompts, after correcting the control.} Each row compares both methods on the same 1,000 questions, and differences are EIR minus the corrected control, calculated before rounding. Intervals are paired bootstrap 95\% intervals with 10,000 resamples and seed 20260806. Holm adjustment covers the six comparisons, and $^{*}$ marks significance after adjustment.}
\label{tab:listfeedback}
\end{table}
\endgroup
\subsection{Revising Dependent Conclusions}
\label{app:repair}\label{app:repairtrans}
\label{app:ladder}

Dependency repair links conclusions to earlier claims they depend on. When a claim changes, extra calls revise affected conclusions and refresh requirements, using the limits in Table~\ref{tab:repro}. The final answer still uses source text. Compared with default EIR, this procedure has lower Answer F1 on five of six comparisons, significantly on HotpotQA and 2WikiMultiHopQA using both models (Table~\ref{tab:ladder}). A second test makes the same repair calls in both versions but applies their proposed changes in only one. Applying changes significantly improves Answer F1 only on MuSiQue using Haiku 4.5 (Table~\ref{tab:repair}).

\begingroup
\footnotesize
\setlength{\tabcolsep}{4pt}
\setlength{\LTpre}{6pt}
\setlength{\LTpost}{6pt}
\setlength{\LTcapwidth}{\textwidth}
\begin{longtable}{llrrrlrr}
\caption{Answer F1 for dependency repair and default EIR. Each row compares both methods on the same 1,000 questions, and differences are Repair minus EIR, calculated before rounding. Holm $p$-values keep the original groups of four comparisons within each benchmark and model, including comparisons not displayed here.}
\label{tab:ladder}\\
\toprule
Model & Benchmark & Repair & EIR & Difference & 95\% CI & Holm $p$ & $\Delta$EM \\
\midrule\endfirsthead
\toprule
Model & Benchmark & Repair & EIR & Difference & 95\% CI & Holm $p$ & $\Delta$EM \\
\midrule\endhead
\bottomrule\endfoot\endlastfoot
Haiku 4.5 & HotpotQA & 68.47 & 72.64 & $-4.17$ & [$-6.02$, $-2.37$] & $<10^{-4}$ & $-1.80$ \\
Haiku 4.5 & 2Wiki & 65.64 & 70.22 & $-4.57$ & [$-6.80$, $-2.30$] & $<10^{-4}$ & $-3.40$ \\
Haiku 4.5 & MuSiQue & 53.32 & 51.98 & $+1.34$ & [$-0.67$, $+3.36$] & 0.1968 & $+1.20$ \\
GPT-4.1 Mini & HotpotQA & 71.08 & 74.05 & $-2.97$ & [$-4.76$, $-1.21$] & 0.0030 & $-4.40$ \\
GPT-4.1 Mini & 2Wiki & 69.83 & 74.89 & $-5.06$ & [$-6.98$, $-3.17$] & $<10^{-4}$ & $-4.60$ \\
GPT-4.1 Mini & MuSiQue & 56.21 & 57.54 & $-1.33$ & [$-3.70$, $+1.03$] & 0.3476 & $-2.10$ \\*
\bottomrule
\end{longtable}
\endgroup

\begingroup
\footnotesize
\setlength{\tabcolsep}{4pt}
\setlength{\LTpre}{6pt}
\setlength{\LTpost}{6pt}
\setlength{\LTcapwidth}{\textwidth}
\begin{longtable}{llrrrlr}
\caption{Answer F1 with proposed repairs applied or left unapplied. Each row compares both methods on the same 1,000 questions, and differences are applied minus unapplied. Holm adjustment covers the six comparisons.}
\label{tab:repair}\\
\toprule
Model & Benchmark & \shortstack[r]{Repairs\\applied} & \shortstack[r]{Detected,\\not applied} & Difference & 95\% CI & Holm $p$ \\
\midrule
\endfirsthead
\toprule
Model & Benchmark & \shortstack[r]{Repairs\\applied} & \shortstack[r]{Detected,\\not applied} & Difference & 95\% CI & Holm $p$ \\
\midrule
\endhead
\bottomrule
\endfoot
\endlastfoot
\multirow{3}{*}{Haiku 4.5}
& HotpotQA & 68.47 & 68.81 & $-0.34$ & [$-2.02$, $+1.26$] & 1.000 \\*
& 2Wiki & 65.64 & 65.37 & $+0.28$ & [$-1.63$, $+2.21$] & 1.000 \\*
& MuSiQue & 53.32 & 51.06 & $+2.26$ & [$+0.67$, $+3.92$] & 0.031 \\
\midrule
\multirow{3}{*}{GPT-4.1 Mini}
& HotpotQA & 71.08 & 71.27 & $-0.20$ & [$-1.86$, $+1.45$] & 1.000 \\*
& 2Wiki & 69.83 & 68.75 & $+1.08$ & [$-0.87$, $+3.01$] & 1.000 \\*
& MuSiQue & 56.21 & 54.34 & $+1.87$ & [$-0.19$, $+3.88$] & 0.379 \\*
\bottomrule
\end{longtable}
\endgroup

\section{Missing Evidence and Additional Retrieval}
\label{app:failure-analysis}

These experiments test whether EIR's errors persist because it did not read the evidence needed. The benchmark labels identify which paragraphs provide that evidence. First, omitted paragraphs are supplied directly. Next, the model chooses additional paragraphs, either for known errors or for all questions.

Reading every labeled paragraph does not guarantee a correct answer. Exact Match also counts differently worded answers as errors when benchmark normalization does not make them identical.

\subsection{Errors After Reading All or Part of the Required Evidence}

Of EIR's 6,000 answers, 2,758 fail Exact Match. EIR had read every paragraph labeled as necessary for 1,510 of these questions, but had missed at least one for 1,248 (45.25\% of the errors, 95\% CI 43.40--47.11\%). Every omitted paragraph was available to read.

\begingroup
\footnotesize
\setlength{\tabcolsep}{4pt}
\setlength{\LTpre}{6pt}
\setlength{\LTpost}{6pt}
\setlength{\LTcapwidth}{\textwidth}
\begin{longtable}{llrrrr}
\caption{\textbf{Evidence read for answers that fail Exact Match.} Rows count EM errors per benchmark and model, divided into those with every annotated supporting paragraph read (Complete) and those missing at least one (Incomplete). Incomplete (\%) divides by EM errors, not by all 1,000 questions.}
\label{tab:failure-phase-a}\\
\toprule
Model & Benchmark & EM errors & Complete & Incomplete & Incomplete (\%) \\
\midrule
\endfirsthead
\toprule
Model & Benchmark & EM errors & Complete & Incomplete & Incomplete (\%) \\
\midrule
\endhead
\bottomrule
\endfoot
\endlastfoot
\multirow{3}{*}{Haiku 4.5}
& HotpotQA & 457 & 286 & 171 & 37.42 \\*
& 2Wiki & 398 & 368 & 30 & 7.54 \\*
& MuSiQue & 593 & 220 & 373 & 62.90 \\
\midrule
\multirow{3}{*}{GPT-4.1 Mini}
& HotpotQA & 412 & 192 & 220 & 53.40 \\*
& 2Wiki & 344 & 265 & 79 & 22.97 \\*
& MuSiQue & 554 & 179 & 375 & 67.69 \\
\midrule
All & All & 2,758 & 1,510 & 1,248 & 45.25 \\*
\bottomrule
\end{longtable}
\endgroup

Of the 1,248 questions with omitted evidence, 1,047 lack one required paragraph, 194 lack two, and seven lack three.

\subsection{Supplying the Missing Supporting Paragraphs}

For these 1,248 errors, the omitted paragraphs are added to $E_T$, giving $E_T^{+}$. The same final answer call receives the question and expanded evidence, without either earlier or reference answers. Benchmark labels determine which paragraphs to add, so this tests the benefit of receiving them, not the ability to find them.

Adding the paragraphs produces 335 exact matches, or 26.84\% (95\% CI 24.46--29.37\%, McNemar $p=2.86\times10^{-101}$). Answer F1 rises from 21.08 to 49.73, a gain of 28.66 points before rounding (95\% CI $[26.38,31.05]$). Table~\ref{tab:failure-c1b} gives the exact-match counts.

Repeating the answer call on unchanged evidence instead produces 17 exact matches, or 1.36\% (95\% CI 0.85--2.17\%), and a 1.22-point Answer F1 gain (95\% CI $[0.63,1.86]$). Thirteen questions become exact under both procedures, 322 only with added evidence, and four only without it. The larger gain with added paragraphs shows that omitted evidence explains some recoverable errors.

\subsection{Selecting More Evidence After a Known Failure}

For the same 1,248 errors, a separate model call selects up to two additional paragraphs from unread titles. It receives only the question, text already read, and available titles. The final answer call then uses the expanded evidence. Benchmark labels select the questions for this test but are hidden from the model.

The model-selected paragraphs yield 174 exact matches, or 13.94\% (95\% CI 12.13--15.97\%, McNemar $p=8.35\times10^{-53}$). Answer F1 rises from 21.08 to 35.86, a gain of 14.78 points (95\% CI $[12.98,16.67]$). Of the 174 matches, 173 follow a successful additional read. This test shows what further retrieval can recover after a known error, but does not test whether the model recognizes its own errors.

\begingroup
\footnotesize
\setlength{\tabcolsep}{4pt}
\setlength{\LTpre}{6pt}
\setlength{\LTpost}{6pt}
\setlength{\LTcapwidth}{\textwidth}
\begin{longtable}{llrrrr}
\caption{\textbf{Exact answers after adding evidence to the same known failures.} Rows are the EM errors missing support ($N$) per benchmark and model. Added support supplies the missing annotated paragraphs, Model-selected lets the model choose at most two more, and Exact (\%) divides Model-selected by $N$.}
\label{tab:failure-c1b}\\
\toprule
Model & Benchmark & \(N\) & Added support & Model-selected & Exact (\%) \\
\midrule
\endfirsthead
\toprule
Model & Benchmark & \(N\) & Added support & Model-selected & Exact (\%) \\
\midrule
\endhead
\bottomrule
\endfoot
\endlastfoot
\multirow{3}{*}{Haiku 4.5}
& HotpotQA & 171 & 46 & 20 & 11.70 \\*
& 2Wiki & 30 & 3 & 3 & 10.00 \\*
& MuSiQue & 373 & 89 & 35 & 9.38 \\
\midrule
\multirow{3}{*}{GPT-4.1 Mini}
& HotpotQA & 220 & 68 & 39 & 17.73 \\*
& 2Wiki & 79 & 18 & 15 & 18.99 \\*
& MuSiQue & 375 & 111 & 62 & 16.53 \\
\midrule
All & All & 1,248 & 335 & 174 & 13.94 \\*
\bottomrule
\end{longtable}
\endgroup

\subsection{Additional Retrieval Without Knowing Which Answers Failed}

The final experiment lets a model decide whether to read more for all 6,000 questions. It sees only the question, retrieved text, and unread titles, then may select up to two additional paragraphs. After a successful read, the final answer call runs again. Otherwise, EIR's original answer is retained.

Across all questions, Answer F1 increases from 66.89 to 69.27, a paired gain of 2.39 points with a 95\% bootstrap interval of $[1.87,2.90]$. Exact Match rises from 54.03\% to 56.17\% ($p=5.14\times10^{-14}$): 211 nonexact answers become exact, while 83 exact answers become nonexact. Support F1 rises from 71.95 to 75.63, and the average fraction of annotated supporting paragraphs read increases from 83.58\% to 91.06\%.

\begingroup
\footnotesize
\setlength{\tabcolsep}{4pt}
\setlength{\LTpre}{6pt}
\setlength{\LTpost}{6pt}
\setlength{\LTcapwidth}{\textwidth}
\begin{longtable}{llrrrrrl}
\caption{\textbf{Additional retrieval applied to every question.} Each row compares both methods on the same 1,000 questions, and More requested (\%) is the percentage for which the model asks to read more. Differences are computed before rounding, and both 2WikiMultiHopQA intervals include zero.}
\label{tab:failure-c2}\\
\toprule
Model & Benchmark & \shortstack{EIR\\Answer F1} & \shortstack{Extended\\Answer F1} & \(\Delta\)F1 & 95\% CI & \(\Delta\)EM & \shortstack{More\\requested (\%)} \\
\midrule
\endfirsthead
\toprule
Model & Benchmark & \shortstack{EIR\\Answer F1} & \shortstack{Extended\\Answer F1} & \(\Delta\)F1 & 95\% CI & \(\Delta\)EM & \shortstack{More\\requested (\%)} \\
\midrule
\endhead
\bottomrule
\endfoot
\endlastfoot
\multirow{3}{*}{Haiku 4.5}
& HotpotQA & 72.64 & 74.54 & \(+1.90\) & [\( +1.06, +2.82 \)] & \(+1.90\) & 17.60 \\*
& 2Wiki & 70.22 & 70.07 & \(-0.15\) & [\( -0.52, +0.17 \)] & \(-0.20\) & 21.00 \\*
& MuSiQue & 51.98 & 54.72 & \(+2.74\) & [\( +1.50, +4.04 \)] & \(+2.30\) & 53.00 \\
\midrule
\multirow{3}{*}{GPT-4.1 Mini}
& HotpotQA & 74.05 & 78.25 & \(+4.20\) & [\( +2.72, +5.72 \)] & \(+3.90\) & 57.50 \\*
& 2Wiki & 74.89 & 75.63 & \(+0.74\) & [\( -0.40, +1.87 \)] & \(+1.00\) & 56.90 \\*
& MuSiQue & 57.54 & 62.43 & \(+4.89\) & [\( +3.01, +6.74 \)] & \(+3.90\) & 85.50 \\
\midrule
All & All & 66.89 & 69.27 & \(+2.39\) & [\( +1.87, +2.90 \)] & \(+2.13\) & 48.58 \\*
\bottomrule
\end{longtable}
\endgroup

The model requests further reading for 2,915 questions (48.58\%), including 1,260 originally answered exactly. Among the requests, 49.19\% concern questions where EIR had missed a required paragraph. The 83 lost exact matches represent 2.56\% of originally exact answers. Thus, additional reading helps overall but also changes some correct answers.

The extension adds 10,841 calls, raising the average from 4.847 to 6.654 per question, a 37.28\% increase. Improving when the model requests more evidence could reduce this extra computation and the loss of correct answers.

\section{Full Pool Evaluation}
\label{app:fullpool}

Full Pool answers once from every paragraph provided with a question, including distractors. It uses the same questions, model versions, final answer instructions, and evidence format as EIR. Paragraphs appear once, in their original order, without shortening. Reference answers and evidence labels are used only for scoring. Settings were fixed before inspecting accuracy: temperature 0.0 and output limits of 500 tokens using GPT-4.1 Mini and 2,048 using Haiku 4.5. Historical settings were recovered from code because complete requests were not saved.

Answer F1 and Exact Match each use a separate group of six tests for Holm correction. Table~\ref{tab:fullpool-paired} gives Answer F1 differences. EIR's two 2WikiMultiHopQA gains also hold for Exact Match. Full Pool's HotpotQA advantage using GPT-4.1 Mini does not reach significance for Exact Match (adjusted $p=0.0630$). Full Pool has higher observed Support F1 in all six comparisons, without a significance test for that metric.

\begin{table}[H]
\centering\small
\begin{tabular*}{\textwidth}{@{\extracolsep{\fill}}llrrrl@{}}
\toprule
Model & Benchmark & EIR & Full Pool & Difference & Paired 95\% CI \\
\midrule
Haiku 4.5 & HotpotQA & 72.64 & 72.11 & $+0.52$ & [$-1.30$, $+2.30$] \\
 & 2Wiki & 70.22 & 64.62 & $+5.60^{*}$ & [$+3.44$, $+7.74$] \\
 & MuSiQue & 51.98 & 53.51 & $-1.53$ & [$-4.04$, $+0.98$] \\
\midrule
GPT-4.1 Mini & HotpotQA & 74.05 & 77.12 & $-3.08^{*}$ & [$-4.88$, $-1.24$] \\
 & 2Wiki & 74.89 & 68.66 & $+6.23^{*}$ & [$+4.30$, $+8.23$] \\
 & MuSiQue & 57.54 & 57.84 & $-0.29$ & [$-2.79$, $+2.18$] \\
\bottomrule
\end{tabular*}
\caption{\textbf{Answer F1 for EIR and Full Pool.} Each row compares both methods on the same 1,000 questions, and differences are EIR minus Full Pool, calculated before rounding. $^{*}$ marks significance after Holm adjustment across the six tests.}
\label{tab:fullpool-paired}
\end{table}

Full Pool uses 6,000 calls, 12.57 million input tokens, and 0.259 million output tokens. EIR uses 29,082 calls, 25.71 million input tokens, and 6.93 million output tokens. At the prices in Table~\ref{tab:reproB}, costs are \$10.01 and \$44.28. All questions remain scored, including ten answer-format failures and three invalid-citation outputs. Eleven connection errors recover through transport retries. There are no terminal provider failures or truncated outputs, and malformed answers are not regenerated.

\section{Full-Corpus Evaluation and Error Analysis}
\label{app:corpus}

\subsection{Questions, Search, and Reading}

This extension uses HotpotQA's 5,233,329 Wikipedia introductory paragraphs from October 2017 and the same 1,000 scored HotpotQA questions using both models. Twenty separate questions per model test the setup. Reusing the scored questions tests transfer to a larger search collection, not performance on new questions.

Search ranks paragraphs using Lucene BM25, which weights matching words by their frequency and adjusts for paragraph length. It uses $k_1=0.9$ for repeated terms and $b=0.4$ for length adjustment.\footnote{\url{https://lucene.apache.org/core/10_5_1/core/org/apache/lucene/search/similarities/BM25Similarity.html}} Titles and text are indexed together with no extra title weight. Words are lowercased, common words removed, and related word forms reduced to a shared stem. Duplicate search terms are merged. Any remaining term can produce a match. Results follow decreasing score, with ties broken by increasing paragraph identifier.

Search returns up to ten titles and identifiers used to request the text, without excerpts, scores, or evidence labels. Only paragraphs found in that trial can be read. Results accumulate, and returned text preserves sentence boundaries. Repeated reads consume another turn and read attempt without duplicating stored evidence.

Each trial permits 12 action turns and six read attempts. Invalid actions consume a turn, and invalid reads also consume a read attempt. Rejections return an explanation. Turn 12 accepts only submission. Any other response ends the trial with an empty submission, after which EIR still answers from the collected text. These rules differ from the main evaluation's six-read budget, so differences between evaluations are descriptive.

\subsection{Model and Baseline Settings}

The model versions are listed in Table~\ref{tab:repro}. All calls use temperature 0.0 and a 2,048-token output limit. Direct is scored on its first valid submission. EIR retains its lists and separate final answer call, without dependency repair or additional calls to refresh requirements.

Agentic SSR uses the rules in Appendix~\ref{app:comparison-adaptation}, but its verifier temperature is 0.0 rather than 0.7. The effect of this change was not tested separately. When grouping corrections for voting, this implementation preserves evidence given as either text or sentence references and distinguishes conflicting text.

Reflexion begins each of up to five trials without earlier source text or search results. Only written reflections carry forward. The external controller stops at an exact match or returns trial five. Answering and reflection prompts contain no reference answers or scores.

\subsection{Scores and Statistical Comparisons}

Tables~\ref{tab:corpus-all} and~\ref{tab:corpus-paired} report scores and within-model comparisons. Oracle-stopped Reflexion includes earlier trials and reflections, so its call total overlaps with the first-trial total. All questions remain scored, including ten interrupted Haiku 4.5 questions. Excluding those ten does not change the conclusion against oracle-stopped Reflexion.

\begin{table}[H]
\centering\footnotesize
\begin{tabular*}{\textwidth}{@{\extracolsep{\fill}}llrrrrr@{}}
\toprule
Model & Method & Answer F1 & EM & Support F1 & Calls & Cost (\$) \\
\midrule
GPT-4.1 Mini & Direct & 48.30 & 27.30 & 50.99 & 6,194 & 3.4639 \\
 & EIR & 61.67 & 46.30 & 55.31 & 7,606 & 5.2582 \\
 & Agentic SSR & 49.14 & 28.70 & 49.25 & 70,292 & 34.4042 \\
 & Reflexion trial 1 & 51.51 & 31.00 & 49.14 & 6,169 & 2.3206 \\
 & Reflexion (oracle) & 56.99 & 38.60 & 54.33 & 27,920 & 15.6130 \\
\midrule
Haiku 4.5 & Direct & 57.07 & 37.70 & 59.19 & 6,329 & 11.3336 \\
 & EIR & 61.35 & 45.60 & 61.44 & 12,025 & 34.3764 \\
 & Agentic SSR & 52.02 & 33.90 & 56.04 & 88,037 & 150.2566 \\
 & Reflexion trial 1 & 50.38 & 32.70 & 55.44 & 8,302 & 12.9287 \\
 & Reflexion (oracle) & 62.21 & 46.60 & 57.06 & 32,555 & 72.6592 \\
\bottomrule
\end{tabular*}
\caption{\textbf{Full-corpus HotpotQA results.} Rows are methods on 1,000 questions per model, with accuracy on a 0--100 scale and calls and costs totaled over all scored questions. Oracle-stopped Reflexion includes every trial and reflection, and costs use the fixed prices in Table~\ref{tab:reproB}.}
\label{tab:corpus-all}
\end{table}

\begin{table}[H]
\centering\footnotesize
\setlength{\tabcolsep}{3pt}
\begin{tabular*}{\textwidth}{@{\extracolsep{\fill}}llrlrr@{}}
\toprule
Model & EIR versus & \shortstack{Answer F1\\difference} & Paired 95\% CI & \shortstack{Answer F1\\Holm $p$} & EM Holm $p$ \\
\midrule
GPT-4.1 Mini & Direct & $+13.38$ & [$+10.91$, $+15.83$] & 0.0010 & $9.938\!\times\!10^{-31}$ \\
 & Agentic SSR & $+12.53$ & [$+10.11$, $+14.96$] & 0.0010 & $9.454\!\times\!10^{-28}$ \\
 & Reflexion trial 1 & $+10.16$ & [$+7.73$, $+12.60$] & 0.0010 & $5.997\!\times\!10^{-21}$ \\
 & Reflexion (oracle) & $+4.68$ & [$+2.01$, $+7.33$] & 0.0024 & $1.019\!\times\!10^{-5}$ \\
\midrule
Haiku 4.5 & Direct & $+4.28$ & [$+2.09$, $+6.43$] & 0.0010 & $9.293\!\times\!10^{-9}$ \\
 & Agentic SSR & $+9.33$ & [$+6.91$, $+11.68$] & 0.0010 & $4.217\!\times\!10^{-16}$ \\
 & Reflexion trial 1 & $+10.97$ & [$+8.59$, $+13.34$] & 0.0010 & $8.664\!\times\!10^{-20}$ \\
 & Reflexion (oracle) & $-0.86$ & [$-3.39$, $+1.65$] & 0.9827 & 1 \\
\bottomrule
\end{tabular*}
\caption{\textbf{Full-corpus paired comparisons, EIR minus the named baseline.} Each row compares both methods on the same 1,000 questions, with Answer F1 differences calculated before rounding and a separate Exact Match test in the last column. Holm $p$-values keep the original groups of five comparisons per metric and model, four of which are shown, and 0.0010 rounds 0.0009999 rather than denoting $p<10^{-4}$.}
\label{tab:corpus-paired}
\end{table}

\subsection{What the Recorded Runs Show}
\label{app:corpus-diagnostics}

The following analyses compare saved answers and retrieved paragraphs for the same questions. Reading every benchmark-labeled paragraph establishes what text was available, not whether the model understood it.

\paragraph{Accuracy Differences on Individual Questions.}
Using Haiku 4.5, EIR scores higher than oracle-stopped Reflexion on 199 questions, lower on 170, and equally on 631. Its mean loss of 66.42 Answer F1 points exceeds its mean gain of 52.43, leaving an overall difference of $-0.86$ points despite more frequent wins. The interval $[-3.39,+1.65]$ and adjusted $p=0.9827$ establish neither higher accuracy nor equivalence, although EIR uses 63.1\% fewer calls. Using GPT-4.1 Mini, 294 higher scores, 176 lower scores, and 530 ties produce the significant $+4.68$-point difference.

Reflexion's later trials add 11.83 observed Answer F1 points using Haiku 4.5 and 5.48 using GPT-4.1 Mini. The largest gains accompany newly read evidence, but reflection, extra retrieval, and stopping at an exact match change together.

\paragraph{Computation After Supporting Evidence Is Read.}
Table~\ref{tab:corpus-after-support} counts calls after all paragraphs labeled as necessary have been read. Later Reflexion trials restart without earlier text, so these counts do not establish that the subsequent calls were unnecessary.

\begin{table}[H]
\centering\footnotesize
\setlength{\tabcolsep}{3pt}
\begin{tabular*}{\textwidth}{@{\extracolsep{\fill}}llrrrrr@{}}
\toprule
Model & Method & Questions & \shortstack{Later actions\\in same trial} & \shortstack{Actions in\\later trials} & Reflections & \shortstack{Final answer\\calls} \\
\midrule
GPT-4.1 Mini & EIR & 420 & 1,004 & --- & --- & 420 \\
 & Reflexion (oracle) & 547 & 846 & 8,442 & 1,299 & --- \\
\midrule
Haiku 4.5 & EIR & 589 & 3,581 & --- & --- & 589 \\
 & Reflexion (oracle) & 542 & 602 & 9,235 & 1,106 & --- \\
\bottomrule
\end{tabular*}
\caption{\textbf{Calls after annotated support is first acquired.} Questions counts runs in which EIR, or at least one Reflexion trial, reads every paragraph labeled as necessary. Counts exclude the read that completes that support, dashes mark call types absent from a method, and Agentic SSR is not covered.}
\label{tab:corpus-after-support}
\end{table}

\paragraph{Effect of EIR's Final Answer Call.}
For the 909 GPT-4.1 Mini runs with an accepted answer during retrieval, final answering raises Answer F1 from 43.62 to 64.83, a gain of 21.20 points before rounding. For 409 Haiku 4.5 runs, it rises from 55.02 to 61.90, a 6.88-point gain. Exact Match improves on 254 questions and is lost on 33 using GPT-4.1 Mini. The corresponding counts using Haiku 4.5 are 55 and 20.

This analysis excludes 91 forced empty submissions using GPT-4.1 Mini and 590 plus one interrupted run using Haiku 4.5. These different subsets should not be used to compare the models' gains. All 2,000 questions remain in the main scores. Their final prompts contain only the question and all saved source sentences, and no final response is marked as truncated. Score changes may reflect answer wording as well as reasoning.

\paragraph{Rejected Actions.}
Actions use JSON, a text format containing named fields. Missing fields, incorrect types, and invalid citations cause rejection (Table~\ref{tab:corpus-rejections}). EIR can extract an action from surrounding prose, while Direct and Reflexion require only JSON. Many rejected Haiku 4.5 Reflexion responses contain extra prose or imitation tool calls. The comparisons therefore include differences in how responses are accepted.

\begin{table}[H]
\centering\footnotesize
\begin{tabular*}{\textwidth}{@{\extracolsep{\fill}}llrrr@{}}
\toprule
Model & Method & Rejected actions & Recorded actions & Rejected (\%) \\
\midrule
GPT-4.1 Mini & EIR & 216 & 6,606 & 3.27 \\
 & Direct & 41 & 6,194 & 0.66 \\
 & Reflexion, all trials & 619 & 25,339 & 2.44 \\
\midrule
Haiku 4.5 & EIR & 2,377 & 11,024 & 21.56 \\
 & Direct & 70 & 6,328 & 1.11 \\
 & Reflexion, all trials & 14,045 & 30,154 & 46.58 \\
\bottomrule
\end{tabular*}
\caption{\textbf{Rejected actions in the full-corpus evaluation.} Rows are methods per model, with denominators counting recorded actions and excluding unfinished requests, reflection, and separate final answering. Reflexion totals include actions from all trials for the same questions.}
\label{tab:corpus-rejections}
\end{table}

Of Haiku 4.5 EIR's 590 forced endings, 323 follow invalid citations, 234 follow incorrectly typed sentence positions or missing paragraph identifiers, and 33 attempt reading or searching on the final turn. Every case still receives the final answer call. No correctly formatted citation to a sentence already read was falsely rejected, although prompts and feedback use several citation formats.


\begin{thebibliography}{21}
\providecommand{\natexlab}[1]{#1}

\bibitem[OpenAI(2025)]{openai2025gpt41}
OpenAI.
\newblock Introducing GPT-4.1 in the API.
\newblock April 14, 2025.
\newblock \url{https://openai.com/index/gpt-4-1/}.

\bibitem[Anthropic(2025)]{anthropic2025haiku45}
Anthropic.
\newblock Introducing Claude Haiku 4.5.
\newblock October 15, 2025.
\newblock \url{https://www.anthropic.com/news/claude-haiku-4-5}.

\bibitem[Asai et al.(2024)]{asai2024selfrag}
Akari Asai, Zeqiu Wu, Yizhong Wang, Avirup Sil, and Hannaneh Hajishirzi.
\newblock Self-RAG: Learning to retrieve, generate, and critique through self-reflection.
\newblock In \emph{International Conference on Learning Representations}, 2024.

\bibitem[Asl et al.(2025)]{asl2025fairrag}
Mohammad Aghajani Asl, Majid Asgari-Bidhendi, and Behrooz Minaei-Bidgoli.
\newblock FAIR-RAG: Faithful adaptive iterative refinement for retrieval-augmented generation.
\newblock \emph{arXiv preprint arXiv:2510.22344}, 2025.

\bibitem[Ho et al.(2020)]{ho2020twowiki}
Xanh Ho, Anh-Khoa Duong Nguyen, Saku Sugawara, and Akiko Aizawa.
\newblock Constructing a multi-hop QA dataset for comprehensive evaluation of reasoning steps.
\newblock In \emph{Proceedings of COLING}, 2020.

\bibitem[Jeong et al.(2024)]{jeong2024adaptiverag}
Soyeong Jeong, Jinheon Baek, Sukmin Cho, Sung Ju Hwang, and Jong C. Park.
\newblock Adaptive-RAG: Learning to adapt retrieval-augmented large language models through question complexity.
\newblock In \emph{Proceedings of NAACL}, pp.\ 7036--7050, 2024.

\bibitem[Liu(2026)]{liu2026sleuth}
Ning Liu.
\newblock Track, rank, crack: Epistemic working memory scales multi-hop reasoning in language agents.
\newblock \emph{arXiv preprint arXiv:2607.12267}, 2026.

\bibitem[Shi et al.(2025)]{shi2025ssr}
Haizhou Shi, Ye Liu, Bo Pang, Zeyu Leo Liu, Hao Wang, Silvio Savarese, Caiming Xiong, Yingbo Zhou, and Semih Yavuz.
\newblock SSR: Socratic self-refine for large language model reasoning.
\newblock \emph{arXiv preprint arXiv:2511.10621}, 2025.

\bibitem[Shinn et al.(2023)]{shinn2023reflexion}
Noah Shinn, Federico Cassano, Ashwin Gopinath, Karthik Narasimhan, and Shunyu Yao.
\newblock Reflexion: Language agents with verbal reinforcement learning.
\newblock In \emph{Advances in Neural Information Processing Systems}, volume~36, 2023.

\bibitem[Trivedi et al.(2022)]{trivedi2022musique}
Harsh Trivedi, Niranjan Balasubramanian, Tushar Khot, and Ashish Sabharwal.
\newblock MuSiQue: Multihop questions via single-hop question composition.
\newblock \emph{Transactions of the Association for Computational Linguistics}, 10:\penalty0 539--554, 2022.

\bibitem[Yang et al.(2018)]{yang2018hotpotqa}
Zhilin Yang, Peng Qi, Saizheng Zhang, Yoshua Bengio, William W. Cohen, Ruslan Salakhutdinov, and Christopher D. Manning.
\newblock HotpotQA: A dataset for diverse, explainable multi-hop question answering.
\newblock In \emph{Proceedings of EMNLP}, 2018.

\bibitem[Yao et al.(2023)]{yao2023react}
Shunyu Yao, Jeffrey Zhao, Dian Yu, Nan Du, Izhak Shafran, Karthik Narasimhan, and Yuan Cao.
\newblock ReAct: Synergizing reasoning and acting in language models.
\newblock In \emph{International Conference on Learning Representations}, 2023.

\end{thebibliography}
\end{document}